\documentclass{article}
\usepackage[margin=1in]{geometry}
\pdfoutput=1

\usepackage[square, numbers, compress, sort]{natbib}

\PassOptionsToPackage{square, numbers, compress, sort}{natbib}

\usepackage[parfill]{parskip}
\usepackage[utf8]{inputenc} 
\usepackage[T1]{fontenc}    

\usepackage[utf8]{inputenc} 
\usepackage[T1]{fontenc}    
\usepackage{hyperref}       
\usepackage{url}            
\usepackage{booktabs}       
\usepackage{amsfonts}       
\usepackage{nicefrac}       
\usepackage{microtype}      
\usepackage{xcolor}         

\usepackage{enumitem}  
\usepackage{macro_math}
\usepackage{booktabs}
\usepackage{multirow}
\usepackage{colortbl} 
\definecolor{Gray}{gray}{0.85}
\definecolor{LightCyan}{rgb}{0.7,1,1}
\definecolor{White}{rgb}{1,1,1}
\newcolumntype{a}{>{\columncolor{Gray}}c}
\newcolumntype{b}{>{\columncolor{LightCyan}}c}
\newcolumntype{n}{>{\columncolor{White}}c}
\usepackage{rotating}
\usepackage{adjustbox}
\usepackage{pifont}
\newcommand{\cmark}{\ding{51}}%
\newcommand{\xmark}{\ding{55}}%
\usepackage{wrapfig}

\usepackage{varwidth}

\usepackage{blindtext}

\makeatletter
\DeclareRobustCommand\onedot{\futurelet\@let@token\@onedot}
\def\@onedot{\ifx\@let@token.\else.\null\fi\xspace}

\def\eg{\emph{e.g}\onedot,} 

\def\ie{\emph{i.e}\onedot} 

\def\cf{\emph{cf}\onedot}

\def\name{contrastive adapting}  
\def\Name{Contrastive adapting}

\def\numberFMs{11}  
\def\numberDatasets{9}

\usepackage[normalem]{ulem}
\useunder{\uline}{\ul}{}

\newcommand{\header}[1]{\textbf{#1.}}

\definecolor{mydarkblue}{rgb}{0,0.08,0.45}
\hypersetup{ %
    pdftitle={},
    pdfauthor={},
    pdfsubject={},
    pdfkeywords={},
    pdfborder=0 0 0,
    pdfpagemode=UseNone,
    colorlinks=true,
    linkcolor=blue,
    citecolor=blue,
    filecolor=blue,
    urlcolor=black,
}

\usepackage{listings}
\definecolor{codegreen}{rgb}{0,0.6,0}
\definecolor{codegray}{rgb}{0.5,0.5,0.5}
\definecolor{codepurple}{rgb}{0.58,0,0.82}
\definecolor{backcolour}{rgb}{0.95,0.95,0.95}

\lstdefinestyle{mystyle}{
    backgroundcolor=\color{backcolour},   
    commentstyle=\color{codegreen},
    keywordstyle=\color{magenta},
    numberstyle=\tiny\color{codegray},
    stringstyle=\color{codepurple},
    basicstyle=\ttfamily\footnotesize,
    breakatwhitespace=false,         
    breaklines=true,                 
    captionpos=b,                    
    keepspaces=true,                 
    numbers=left,                    
    numbersep=5pt,                  
    showspaces=false,                
    showstringspaces=false,
    showtabs=false,                  
    tabsize=2
}
\title{Contrastive Adapters for Foundation Model Group Robustness}

\date{}

\author{%
  Michael Zhang\thanks{Corresponding author}\; and Christopher R\'{e} \\
  Stanford University\\
  \texttt{\{mzhang,chrismre\}@cs.stanford.edu} \\
}

\begin{document}


\maketitle

\begin{abstract}
\noindent
While large pretrained foundation models (FMs) have shown remarkable zero-shot classification robustness to dataset-level distribution shifts, their robustness to subpopulation or group shifts is relatively underexplored. We study this problem, and find that FMs such as CLIP may not be robust to various group shifts. Across 9 robustness benchmarks, zero-shot classification with their embeddings results in gaps of up to 80.7 percentage points (pp) between average and worst-group accuracy. Unfortunately, existing methods to improve robustness require retraining, which can be prohibitively expensive on large foundation models. We also find that efficient ways to improve model inference (\eg{} via adapters, lightweight networks with FM embeddings as inputs) do not consistently improve and can sometimes \emph{hurt} group robustness compared to zero-shot (\eg{} increasing the accuracy gap by 50.1 pp on CelebA). We thus develop an adapter training strategy to effectively and efficiently improve FM group robustness. Our motivating observation is that while poor robustness results from groups in the same class being embedded far apart in the foundation model ``embedding space,'' standard adapter training may not bring these points closer together. We thus propose contrastive adapting, which trains adapters with contrastive learning to bring sample embeddings close to both their ground-truth class embeddings \emph{and other sample} embeddings in the same class. Across the 9 benchmarks, our approach consistently improves group robustness, raising worst-group accuracy by 8.5 to 56.0 pp over zero-shot. Our approach is also efficient, doing so without any FM finetuning and only a fixed set of frozen FM embeddings. On benchmarks such as Waterbirds and CelebA, this leads to worst-group accuracy comparable to state-of-the-art methods that retrain entire models, while only training $\leq$1\% of the model parameters.

\end{abstract}

\section{Introduction}

Foundation models (FMs)---large pretrained models trained on massive datasets---offer an exciting new paradigm for deep learning. Recent works have shown that without any finetuning, foundation models can generalize well to various datasets \cite{brown2020language, radford2021learning, jia2021scaling, singh2021flava} and exhibit impressive robustness to certain distribution shifts \citep{kumar2022fine, wortsman2021robust}. Under this zero-shot paradigm, 
practitioners can avoid training task-specific models, and instead 
use FM embeddings for efficient and effective inference. 

However, 
an underexplored question is how robust this zero-shot inference is to ``group shifts,'' distribution shifts between subpopulations or meaningful groups in data. Prior works have established that \emph{group robustness}---\ie{} performing well on all groups---is a fundamental and real-world challenge for modern deep learning \cite{sagawa2020investigation, sohoni2020no, nam2020learning, beery2018recognition, buolamwini2018gender, oakden2020hidden, koh2021wilds}. Yet most prior foundation model evaluations focus on overall or average performance \citep{radford2021learning, kumar2022fine, wortsman2021robust}; few works consider their accuracy across groups.


This work thus further studies the group robustness of foundation models for zero-shot classification.
We first motivate this problem by showing that foundation models can have poor zero-shot group robustness. Evaluating \numberFMs{} foundation models across \numberDatasets{} robustness benchmarks, we find they achieve up to an 80.7 percentage point (pp) gap between average and worst group accuracy, where a CLIP ResNet-50~\cite{radford2021learning} only classifies 6.0\% of worst-group samples correctly on the CelebA dataset~\cite{liu2015deep, sagawa2019distributionally}. 

Given this observed weakness, we aim to improve foundation model group robustness.
This poses several challenges and open questions to answer. First, while improving group robustness in machine learning is well-studied, existing robustness methods require retraining one (and often more than one) entire models \citep{sagawa2019distributionally, nam2020learning, sohoni2020no, creager2021environment, liu2021just, Ahmed2021SystematicGW, taghanaki2021robust, zhang2022correct, kirichenko2022last}. This can be prohibitively expensive for foundation models due to their size and scale, raising the question of whether we can make these models more robust without any retraining or finetuning. Second, for zero-shot classification, many practitioners may also only access foundation model outputs or embeddings (\eg{} via  APIs\footnote{\href{https://beta.openai.com/docs/introduction.}{https://beta.openai.com/docs/introduction.}, \href{https://studio.ai21.com/docs/}{https://studio.ai21.com/docs/}, \href{https://docs.cohere.ai/}{https://docs.cohere.ai/}}). To improve robustness, ideal solutions should only require pretrained FM embeddings. However, these same embeddings lead to poor zero-shot robustness, raising the question of whether they even encode the information needed to classify all groups correctly.

Motivated by these challenges and questions, we study effective \emph{and} efficient solutions for better FM group robustness. As a baseline, we first find that while efficient methods to improve FM inference---such as training linear probes (linear classifiers)~\cite{radford2021learning, kumar2022fine} and adapters (small bottleneck MLPs)~\cite{pmlr-v97-houlsby19a, gao2021clip} on top of FM embeddings---can improve group robustness over zero-shot (reducing the gap by up to 50.2 pp on representative benchmarks), they fail to do so consistently, and 
can \emph{hurt} robustness. They reduce worst-group accuracy by up to 37.9 pp, and 
increase the accuracy gap by up to 74.9 pp. To reason about this inconsistency, we note that 
poor zero-shot robustness results when 
FMs embed same-class samples in different groups ``far apart'' in embedding space. While adapter training achieves higher robustness than linear probing, we find settings where it still fails to close this distance, \eg{} if training data is group-imbalanced. 

\begin{figure}[!t]
  \vspace{-0.5cm}
  \centering
  \includegraphics[width=1\textwidth]{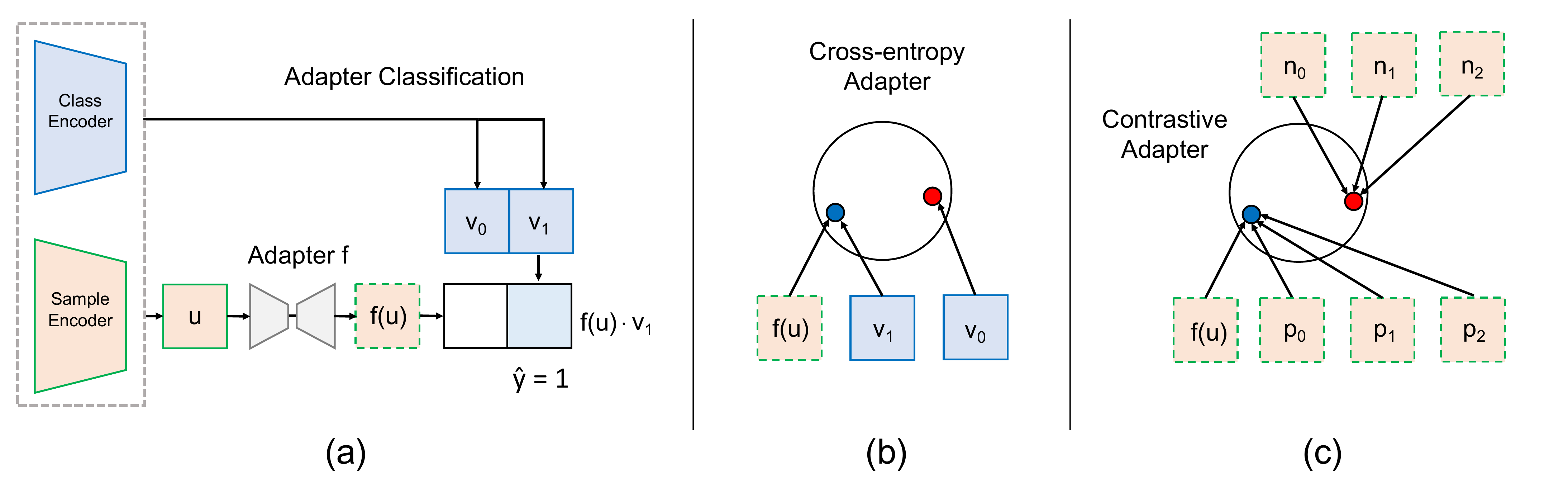}
  \vspace{-0.6cm}
  \caption{(a) Adapter classification with FM embeddings. Adapters learn transformations to align sample embeddings to ground-truth class embeddings. (b) Cross-entropy loss encourages alignment between class embeddings~\cite{gao2021clip}. (c) \Name{} adds alignment between sample embeddings. }
  \label{fig:cos_sim_motivation}
\end{figure}

To then handle the setting where baselines hurt robustness, and consistently improve group robustness over zero-shot, we propose \emph{\name{}}, a simple adapter training method that places greater emphasis on bringing these initially ``far apart'' points together.
For each task, we first use foundation models to compute embeddings for each training sample and class. We then train adapters on these embeddings. Like prior work~\cite{gao2021clip}, these adapters take sample embeddings as inputs, and output transformed embeddings with greater cosine similarity to their ground-truth class embeddings. 
However, the key difference is that 
\name{} also applies a supervised contrastive loss over other \emph{sample} embeddings. Specifically, we provide a way to ``pull together'' far apart sample embeddings in the same class, and ``push apart'' nearby sample embeddings in different classes. 


In our experiments, we validate that \name{} effectively and efficiently improves FM group robustness. First, across all 9 robustness benchmarks, we find \name{} consistently improves worst-group accuracy over zero-shot (by 8.5 to 56.0 pp), using no training group labels and only training MLPs with 0.1\% to 0.3\% of the original FM parameters. Then, on a representative set of benchmarks with various group shifts and training data group sizes, we find \name{} can substantially improve group robustness over prior adapter training strategies, and outperforms other approaches that only use fixed FM embeddings (improving worst-group accuracy by 12.4 pp and average accuracy by 0.6 pp than the next best robustness method on average with CLIP ResNet-50 embeddings).
Finally, beyond just improving FM robustness, we find \name{} also achieves effective and efficient group robust classification in general. We achieve near state-of-the-art (SoTA) or SoTA worst-group accuracy on popular robustness benchmarks with only 1.0\% of the trainable parameters of the SoTA models (\eg{} improving 0.2 pp over the prior SoTA~\cite{nam2022spread} on CelebA~\cite{liu2015deep} by training a single adapter instead of a ResNet-50).




We summarize our contributions as follows:
\begin{itemize}[leftmargin=*]
    \item We demonstrate that zero-shot classification with existing foundation models may not be group robust across a variety of popular group robustness benchmarks. 
    \item We propose a contrastive adapter training strategy that efficiently improves foundation model group robustness \emph{without any} finetuning of the original model weights. 
    \item We find that \name{} more effectively improves foundation model group robustness over similarly efficient alternatives that do not require finetuning, and more efficiently achieves comparable group robustness to state-of-the-art approaches outside the foundation model regime on standard benchmarks. 
\end{itemize}
Our results suggest that while pretrained foundation models ``out-of-the-box'' may not classify all groups correctly, the information to classify groups frequently \emph{is} in their pretrained embeddings. Rather than having to train additional separate models, we may just need the right methods to extract this information from pretrained foundation model embeddings.

\section{Related Work}
\vspace{-0.25cm}
\label{sec:related_work}
Our work builds on (i) methods to improve group robustness, and (ii) methods to improve foundation model inference without accessing or finetuning their original weights. We briefly describe these works here, and include an expanded discussion in Appendix~\ref{appendix:extended_related_work}.

\header{Improving group robustness} Many works aim to improve group robustness. If training group labels are known, prior methods often balance group sizes during training, via sample balancing \cite{he2009learning,cui2019class, idrissi2021simple, kirichenko2022last}, importance weighting \cite{shimodaira2000improving, byrd2019effect}, or robust optimization \cite{sagawa2019distributionally, arjovsky2019invariant}. In this work we do not assume training group labels. Without training group labels, a common approach first trains a model with empirical risk minimization (ERM), before using this model's predictions to infer groups. Methods then train a second robust model with sample balancing \cite{nam2020learning, liu2021just} or robust optimization \cite{sohoni2020no, creager2021environment, nam2022spread} using inferred group labels, or representation learning to learn similar representations for groups in the same class \cite{zhang2022correct}. While effective at improving group robustness, these solutions require training one (and often more than one) models. This can make applying them to foundation models prohibitively expensive.

\header{Improving foundation model inference efficiently} Other prior works aim to improve foundation model downstream performance, without having to finetune or update original model weights. \emph{Prompt tuning} optimizes the inputs of a FM while keeping the original model weights frozen. Optimizing either text \cite{li2021prefix, zhou2021learning, zhou2022conditional, levine2022standing} or image \cite{bahng2022visual, yao2021cpt} inputs can improve a frozen foundation model's downstream task accuracy. However, doing so can require multiple passes through the foundation model, which may become expensive in certain situations (\eg{} interacting with a foundation model via a commercial API).
Another paradigm adds small trainable parameters to the original model, either within its layers or on top of its embeddings. These include linear probes (linear classifiers) \cite{radford2021learning} and adapters (small bottleneck MLPs) \cite{pmlr-v97-houlsby19a, Rebuffi17, pfeiffer-etal-2021-adapterfusion, pfeiffer-etal-2020-adapterhub}. Recently, \citet{kumar2022fine, wortsman2021robust} propose methods that use linear probes to improve robustness to dataset-level out-of-distribution shifts. 
\citet{gao2021clip} propose to train single adapters on top of FM embeddings to improve average downstream task accuracy. In contrast, we focus on \emph{group shifts} that occur within a dataset. We also show that prior adapter training can hurt group robustness, and propose alternatives to consistently improve group robustness.

\section{Problem}
\label{sec:problem}

In Section~\ref{sec:group_robustness_problem}, we first describe the group robustness problem setting. In Section~\ref{sec:foundation_model_group_robustness_problem}, we illustrate this problem with foundation models. We show that zero-shot classification with foundation models, and existing baseline approaches to improve downstream inference, can result in poor group robustness.

\subsection{Preliminaries: group robustness and task setup}
\vspace{-0.125cm}
\label{sec:group_robustness_problem}
We emphasize robustness to distribution shifts between groups in this work. For setup, we follow prior works \cite{sagawa2019distributionally, sohoni2020no, liu2021just, koh2021wilds} that alternatively describe the phenomenon as \emph{hidden stratification} \cite{sohoni2020no} or \emph{subpopulation shift} \cite{koh2021wilds}. For some task, we have $N$ samples $\{(x_i, y_i, g_i)\}_{i=1}^N$, with sample features or inputs $x_i \in \cX$, class labels $y_i \in \cY$, and group labels $g_i \in \cG$. Let $C = |\cY|$ be the number of classes. We use $g_i$ to indicate the group that each sample belongs in, but do not observe group labels during training. Distribution shifts may occur between samples in different groups but the same class. 

Every sample $(x_i, y_i, g_i)$ is drawn from some unknown joint distribution $P$. Let $P_g$ be the specific distribution conditioned on $g$ for any $g \in \mathcal{G}$. For classification loss $\ell : \mathcal{Y} \times \mathcal{Y} \mapsto \mathbb{R}$ and classifier $f_\theta: \mathcal{X} \mapsto \mathcal{Y}$, we want $f_\theta$ to be
accurate, \ie{} achieving low average error:
\begin{equation}
    \mathcal{L}_\text{avg}(f_\theta) \define \mathbb{E}_{(x, y, g) \sim P} [\ell(f_\theta(x), y)]
\label{eq:average_loss}
\end{equation}
and \emph{group robust}, \ie{}, achieving a small gap between its average error and its worst-group error:
\begin{equation}
    \mathcal{L}_\text{wg}(f_\theta) \define \max_{g \in \cG}\; \mathbb{E}_{(x, y, g) \sim P_g} [\ell(f_\theta(x), y)]
\label{eq:worst_group_loss}
\end{equation}
Different from domain generalization or out-of-distribution (OOD) evaluation, we observe each group in training, validation, and test splits. However, standard training via empirical risk minimization (ERM) can still lead to poor test set group robustness because training groups may be imbalanced \cite{sagawa2019distributionally, sohoni2020no, zhang2022correct}. 
Here, foundation models are \emph{not trained} on these dataset-specific training sets, so it is not obvious that these models will exhibit the same non-robust behavior as ERM-trained models in prior work~\cite{sagawa2019distributionally, sohoni2020no, zhang2022correct}. However, we show that zero-shot classification with foundation models such as CLIP~\cite{radford2021learning} and GPT-Neo~\cite{gpt-neo} can still result in poor group robustness. 

\begin{figure}[t]
  \vspace{-0.25cm}
  \centering
  \includegraphics[width=1\textwidth]{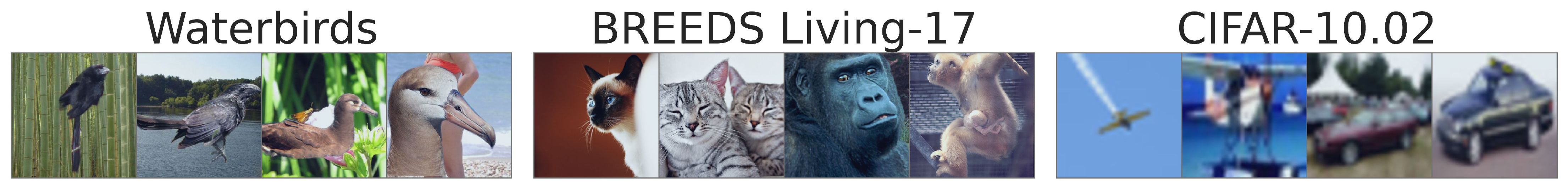}
  \vspace{-0.75cm}
  \caption{Samples of different group shifts for robust evaluation (2 classes, 2 groups per class shown).}
  \label{fig:example_group_shifts}
\end{figure}

\subsection{Empirical findings of poor foundation model group robustness}
\label{sec:foundation_model_group_robustness_problem}
To motivate the rest of this work, we now demonstrate the group robustness problem with foundation models. We first describe  different natural group shifts for evaluation. We next detail primary baseline approaches. We finally summarize our findings after evaluating these baselines on \numberFMs{} popular foundation models across \numberDatasets{} standard group robustness benchmarks used in prior work \cite{sagawa2019distributionally, santurkar2020breeds, recht2018cifar, lu2020harder, koh2021wilds}. In this section, we present four representative scenarios based on training data assumptions and group robustness outcome, deferring the complete results to Appendix~\ref{appendix:extended_zeroshot_evaluation}. Critically, we find that zero-shot classification with foundation models may result in poor group robustness. We also find that baseline methods to improve downstream transfer do not consistently improve group robustness, sometimes decreasing worst-group accuracy.

\header{Dataset group shifts} We benchmark methods on the following sources of group shift (Figure~\ref{fig:example_group_shifts}):
\begin{itemize}[leftmargin=*]
    \item \textbf{Spurious confounders.} We evaluate across groups distinguished by ground-truth class and spurious confounders---input features predictive for some, but not all groups in a class. For example, in Waterbirds~\cite{WelinderEtal2010, sagawa2019distributionally}, a water background is a confounder for the \texttt{waterbirds} class.
    \item \textbf{Subclass variance.} We evaluate across groups which are different fine-grained subclasses. For example, in BREEDS Living-17~\cite{santurkar2020breeds}, the \texttt{ape} class includes images of gibbons and gorillas. 
    
    \item \textbf{Data source variance.} We evaluate across groups which are the same class but sourced from different datasets. For example, we set up the CIFAR-10.02 dataset by combining CIFAR-10~\cite{Krizhevsky2009LearningML} and CIFAR-10.2~\cite{lu2020harder}. The \texttt{airplanes} class contains samples from both datasets.
\end{itemize}

\header{Baseline methods} To evaluate foundation model group robustness, we consider the following baseline methods. Following prior work~\cite{radford2021learning, jia2021scaling, furst2021cloob, mu2021slip, li2021supervision}, using a foundation model for all approaches we first compute sample input embeddings (which we denote as sample embeddings) for all samples we wish to classify, and $C$ class label embeddings (which we denote as class embeddings) for all $C$ classes. With foundation model embedding dimension $D$, let $u_n \in \mathbb{R}^D$ be a sample embedding and $c_n \in \mathbb{R}^D$ be a class embedding.

\begin{itemize}[leftmargin=*]
    \item \textbf{Zero-shot classification} \cite{radford2021learning}:
    We classify each sample via the nearest class embedding to its sample embedding $u_n$. 
    Specifically, 
    we compute the class-wise probabilities for each sample $x_n$ as
    \begin{equation}
    f_\theta(x_n; \tau) = \text{softmax}(\hat{W}^\top \hat{u}_n / \tau)
    \end{equation}
    where $\hat{u}_n = u_n / \lVert u_n \rVert$ is the ($\ell^2$-)normalized sample embedding of $x_n$, $\hat{W} \in \mathbb{R}^{D \times C}$ is a matrix whose columns are the normalized class embeddings $\{\hat{v}_c\}_{c=1}^C$, and $\tau$ is a temperature parameter.
    
    As standard, for class embeddings we convert each class name to a natural language prompt, \eg{} ``photo of a \texttt{[class name]}'', and feed the tokenized prompt to a foundation model's text encoder.
    
    \item \textbf{Linear Probe} \cite{radford2021learning, wortsman2021robust}:
    We train a linear classifier on top of training data sample embeddings. Specifically, with classifier $f_\theta(u) = W^\top u$, we update the weights $W \in \mathbb{R}^{D \times C}$ with a cross-entropy loss applied over training data sample embeddings $\{u_n\}_{n=1}^N$ and labels $\{y_n\}_{n=1}^N$.
    
    \item \textbf{Adapter} \cite{gao2021clip, Rebuffi17}: 
    We train a single 2-layer bottleneck multilayer perception (MLP) to output transformed sample embeddings, which we use instead of the original sample embeddings to classify with in the zero-shot procedure above. 
    Specifically, with adapter hidden-layer dimension $H$, ReLU activation function $\sigma$, and adapter weights $\phi =[W_1, W_2]$---where $W_1 \in \mathbb{R}^{D \times H}$ is a linear down-projection and $W_2 \in \mathbb{R}^{H \times D}$ a linear up-projection---we compute ``adapted'' embeddings
    \begin{equation}
        f_\phi(u) = W_2^\top \sigma \left( W_1^\top u \right)
    \end{equation}
    We classify samples with the zero-shot class matrix $\hat{W}$, temperature $\tau$, and normalized adapted embeddings $\hat{f}_\phi(u) = f_\theta(u) / \lVert f_\phi(u) \rVert$. The final outputs are given by \(f_\theta(u; \hat{W}, \tau) = \hat{W}^\top\hat{f}_\phi(u) / \tau \). Like with linear probes, we update $\phi$ with a cross-entropy loss using training data labels $\{y_n\}_{n=1}^N$.
\end{itemize}

For evaluation, we train both linear probes and adapters with standard empirical risk minimization (ERM), which aims to minimize the empirical risk:
\(
    \hat{\cL}(f_\theta) = \frac{1}{N} \sum_{n =1}^N \ell(f_\theta(u_n), y_n)
\).

\begin{table}[!t]
\caption{Baseline worst-group (WG) and average (Avg) accuracies with zero-shot classification, linear probes, and adapters. Best metric \textbf{in bold}. While training linear probes and adapters can improve group robustness (reducing the worst-group versus average accuracy gap by 57.4 pp on BREEDS Living-17), it can also result in poorer robustness (in {\color[HTML]{FD6864} red}), increasing the gap by 74.9 pp on CelebA.}
\label{tab:representative_outcomes_metrics}
\centering
\begin{adjustbox}{width=\textwidth}
\begin{tabular}{@{}lccbccbccbccb@{}}
\toprule
Method       & \multicolumn{3}{c}{Waterbirds}                     & \multicolumn{3}{c}{CelebA}                         & \multicolumn{3}{c}{BREEDS Living-17}                  & \multicolumn{3}{c}{CIFAR-10.02}                    \\ \cmidrule(l){2-4} \cmidrule(l){5-7} \cmidrule(l){8-10} \cmidrule(l){11-13}  
Accuracy (\%)     & WG            & Avg  & Gap                         & WG            & Avg  & Gap                         & WG            & Avg  & Gap                         & WG            & Avg  & Gap                         \\ \midrule
Zero-shot    & 36.6          & 92.2 & 55.6                        & \textbf{74.0} & 81.9 & \textbf{7.9}                & 6.0           & 86.7 & 80.7                        & 39.1          & 69.9 & 30.8                        \\
Linear Probe & {\color[HTML]{FD6864} 7.9    }       & 93.5 & {\color[HTML]{FD6864} 85.6} & 11.9          & 94.7 & {\color[HTML]{FD6864} 82.8} & 53.3          & 90.8 &  37.5 & 51.3          & 77.7 &  26.4 \\
Adapter      & \textbf{60.8} & 96.0 & \textbf{35.2}               & 36.1          & 94.2 & {\color[HTML]{FD6864} 58.1} & \textbf{70.7} & 94.0 & \textbf{23.3}               & \textbf{68.8} & 86.0 & \textbf{17.2}               \\ \bottomrule
\end{tabular}
\end{adjustbox}
\end{table}

\begin{table}[!t]
\tabcolsep=0.15cm
\caption{Representative outcomes for improving group robustness.}
\label{tab:representative_outcomes}
\centering
\begin{tabular}{@{}lccccccc@{}}
\toprule
     & &  \multicolumn{3}{c}{Class-wise Group Size} & \multicolumn{2}{c}{Improved Group Robustness?} \\ \cmidrule(l){3-5} \cmidrule(l){6-7} 
Example Dataset &  Group Shift      & Largest       & Smallest  & Balanced?      & Linear Probe          & Adapter               \\ \midrule
Waterbirds    & Confounder & 1057          & 56    & \xmark         & \xmark & \cmark \\
CelebA        & Confounder & 22880         & 1387     & \xmark        & \xmark & \xmark \\
BREEDS Living-17 & Subclass & 1076          & 1009    & \cmark       & \cmark & \cmark \\
CIFAR-10.02   & Data source & 4039          & 431      & \xmark      & \cmark & \cmark \\ \bottomrule
\end{tabular}
\end{table}

\header{Discussion and representative outcomes} In Table~\ref{tab:representative_outcomes_metrics}, we report worst-group and average accuracies along with their corresponding gaps on four representative group robustness datasets, using zero-shot classification, linear probes, and adapters on CLIP ResNet-50 embeddings. 
We select datasets to report based on training data setup and group robustness outcome, where we find that (i) the relative group size ratios, (ii) the type of group shift, and (iii) the choice of adapter or linear probe influences group robustness improvements. We note dataset descriptive characteristics and outcomes in Table~\ref{tab:representative_outcomes}, and summarize three main takeaways below. Appendix~\ref{appendix:extended_zeroshot_evaluation} contains results for all datasets and models.

\begin{enumerate}[leftmargin=*]
    \item[1] \textbf{Foundation model zero-shot classification may not be group robust}: Across datasets, we find that zero-shot classification with CLIP ResNet-50 embeddings can achieve 7.9 to 80.7 pp gaps between worst-group and average accuracy. Worryingly, poor group robustness is accompanied by high \emph{average} error (from 69.9\% to 92.9\%), the usual metric for evaluating zero-shot classification. This further supports the importance of improving group robustness.
    
    \item[2] \textbf{Efficient baselines do not consistently improve robustness}: We find that while previously proposed linear probes and adapters are efficient ways to improve accuracy on downstream tasks, these benefits do not consistently carry over to improving group robustness.
    \begin{itemize}
        \item When training data is balanced, both linear probes and adapters can substantially improve group robustness and worst-group accuracy (reducing the robustness gap by 43.2 and 54.7 pp respectively on BREEDS Living-17). However, when minority groups are rare, in some instances, approaches can hurt group robustness. 
        On CelebA, adapters and linear probes increase the gap by 50.2 and 74.9 pp, and reduce worst-group accuracy by 37.9 and 62.1 pp.
    \end{itemize}
    
    \item[3] \textbf{We can improve group robustness with only foundation model embeddings}: Our positive results on BREEDS Living-17, CIFAR-10.02, and Waterbirds suggest that poor zero-shot classification may not be because sample embeddings lack the information required to classify groups correctly. Rather, we may just require the right strategies to ``extract'' this latent information from the pretrained embeddings, and learn how to better classify by this information. 

\end{enumerate}
Altogether, takeaways 1 and 2 motivate the need for methods to effectively improve robustness in the foundation model setting. Takeaway 3 suggests we can make progress on this problem by just learning how to better classify fixed foundation model embeddings.

\section{Method}
\label{sec:method}
Having established the group robustness problem in Section~\ref{sec:problem}, we now propose a simple contrastive adapter training strategy to improve group robustness. In Section~\ref{sec:limitations_embedding_metric}, we setup our approach by identifying possible sources of limitation with standard adapter training.  
In Section~\ref{sec:method_algorithm}, we then use these insights to propose a simple yet effective approach that counteracts these limitations. 

\subsection{Understanding prior limitations via embedding metrics}
\label{sec:limitations_embedding_metric}

To guide a first-step strategy for improving robustness, we first outline high-level reasoning for why zero-shot and ERM-trained adapters fail to classify groups correctly. 
Recall that a key property of group robust classification is that all sample embeddings belonging to the same class should embed closer to their ground-truth class embedding than any other class embedding. 
If zero-shot classification for a specific class is accurate on average but not group robust, then in the pretrained foundation model embedding space there exists groups that embed ``close'' to their ground-truth class embedding, and groups in the same class that embed ``far away'' (measured via cosine similarity).
One way to interpret training adapters with FM embeddings is that it aims to bring these initially far apart sample embeddings closer to their ground-truth class embedding. 
Restating the sample cross-entropy loss with adapters makes this clear as an InfoNCE loss \cite{oord2018representation, chen2020simple}:
\begin{equation}
    \ell(f_\theta(u), y) = - \log \frac{\exp(\hat{f}_\theta(u)^\top \hat{v} / \tau)}{\sum_{c=1}^C \exp(\hat{f}_\theta(u)^\top \hat{v}_c / \tau)}
    \label{eq:cross_entropy_adapter}
\end{equation}
 with sample embedding $u$ as an anchor, class embedding $v$ of ground-truth $y$ as a single positive, and the other $C-1$ class embeddings as negatives. Ideally, via ERM adapters we thus learn transformations bringing zero-shot incorrect anchors closer to their class embedding positives. 
 
 
However, in Section~\ref{sec:problem} we found this loss works in some scenarios but not others. Intuitively, Eq.~\ref{eq:cross_entropy_adapter} can fail to bring samples closer to their correct class embedding (\eg{} on CelebA). To find additional ways to bring points together, 
we hypothesize that 
poor robustness also accompanies poor similarity between 
%
%
\emph{sample embeddings} from different groups but the same class. We verify this in Figure~\ref{fig:cos_sim_motivation} by empirically measuring the average pairwise cosine similarity and group alignment loss $\cL_\text{align}$~\cite{zhang2022correct}---which measures the pairwise Euclidean distance---between sample embeddings in the same class but different groups. We compare these metrics with embeddings computed with trained adapters and the initial foundation model embeddings, and find that higher worst-group accuracy corresponds to higher cosine similarity and lower alignment loss between groups in the same class. 


\begin{figure}[!h]
  \vspace{-0.25cm}
  \centering
  \includegraphics[width=1\textwidth]{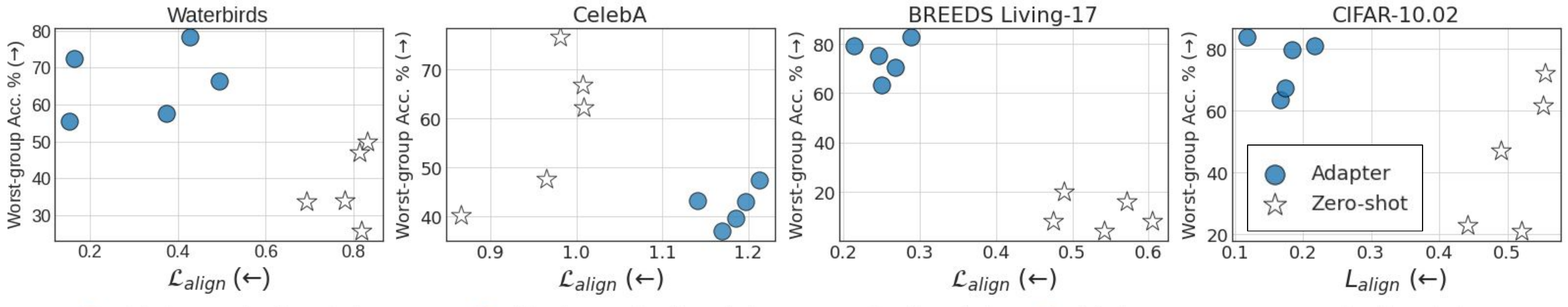}
  \vspace{-0.5cm}
  \label{fig:loss_align_motivation}
\end{figure}
\begin{figure}[!h]
  \vspace{-0.5cm}
  \centering
  \includegraphics[width=1\textwidth]{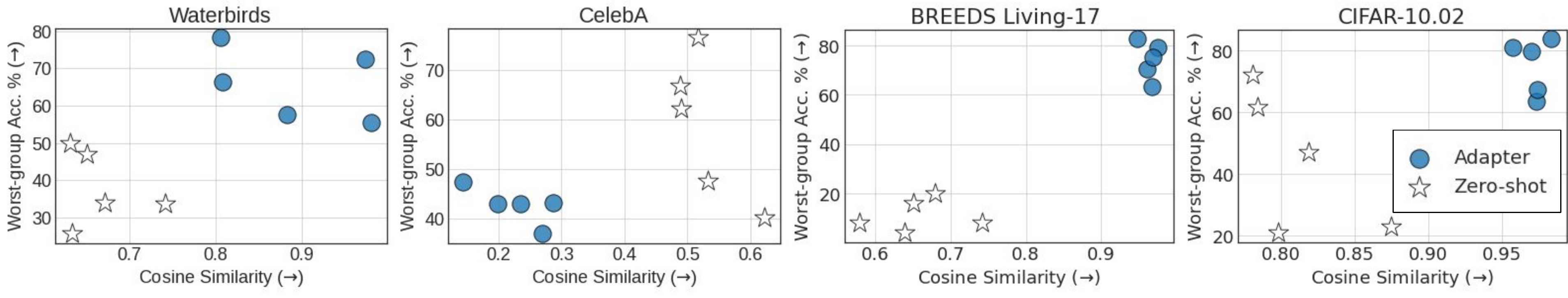}
  \vspace{-0.5cm}
  \caption{Across CLIP model architectures, cosine similarity and alignment loss between groups of the same class tracks worst-group error. Notably, training ERM adapters may fail to move these metrics in the desired direction, which corresponds with poorer robustness (\eg{} on CelebA).}
  \label{fig:cos_sim_motivation}
  \vspace{-0.25cm}
\end{figure}

\vspace{-0.125cm}
\subsection{Approach: Contrastive Adapting}
\label{sec:method_algorithm}
\vspace{-0.125cm}

To improve robustness, we therefore propose to more effectively bring far away samples together by introducing greater training signal via other \emph{sample} embeddings. Instead of limiting ourselves to a single class embedding positive and a limited set of $C - 1$ negatives, we expand our positives by including \emph{sample} embeddings for points in the same class far away from the anchors among pretrained embeddings (\eg{} likely in different groups). We expand our negatives with sample embeddings from different classes. Following prior work~\cite{zhang2022correct, ge2021robust} that finds sampling \emph{hard negatives} beneficial for robust contrastive learning, we also use the computed foundation model sample embeddings to sample negatives from points nearest to the anchors but in different classes. As the number of training data points $N$ is often much larger than the number of classes $C$, these choices are further supported by prior work suggesting more positives and negatives are beneficial for contrastive learning~\cite{Khosla2020Supcon, Robinson2020ContrastiveLW}.
In practice, \name{} is simple to implement with three components: 
\begin{itemize}[leftmargin=*]
    \item \textbf{Foundation model embedding and prediction}: We compute FM embeddings over labeled training data. To guide sampling, we collect zero-shot predictions over this data.
    
    \item \textbf{Contrastive sampling}: For each class, we identify an ``anchor'' sample embedding $u \in \mathcal{U}$ that zero-shot predicts incorrectly, and $P$ ``positive'' sample embeddings $\mathcal{P}(u)$ that zero-shot classifies correctly. We do this as a heuristic for finding samples ``far apart'' in the FM embedding space, so pushing them together improves robustness over zero-shot. We also identify $M$ hard ``negative'' sample embeddings $\mathcal{M}(u)$ by computing the nearest neighbors to the anchors in different classes, using cosine similarity between the sample embeddings.
    
    \item \textbf{Training objective}: We use a supervised contrastive loss~\cite{Khosla2020Supcon} with the sample embeddings, \ie{}
    \begin{equation}
       \ell_\text{con}^\text{sup}(f_\theta(u)) = \frac{-1}{P} \sum_{p \in \mathcal{P}(u)} \log \frac{\exp(\hat{f}_\theta(u)^\top \hat{f}_\theta(p) / \tau)}{\exp(\hat{f}_\theta(u)^\top \hat{f}_\theta(p) / \tau) + 
        \sum_{m \in \mathcal{M}(u)} \exp(\hat{f}_\theta(u)^\top \hat{f}_\theta(m) / \tau)} 
        \label{eq:contrastive_adapter}
    \end{equation}
    We also use a standard cross-entropy loss over minibatches of sample embeddings and their class embeddings. This aims to keep adapted embeddings close to their ground-truth class embeddings. To avoid undoing Eq.~\ref{eq:contrastive_adapter} and push ``far away'' points closer to their ground-truth class embeddings, we upsample the number of zero-shot incorrect samples to equal the number of zero-shot correct samples in each minibatch. We thus use contrastive supervision from class and sample embeddings.
\end{itemize}

\header{Robust generalization with adapters} The contrastive loss in Equation~\ref{eq:contrastive_adapter} is also supported by recent results suggesting that minimizing the class-wise alignment loss $\mathcal{L}_\text{align}$ helps bound the worst-group versus average error gap for that class (\cf Thm~3.1, \citet{zhang2022correct}). The bound however scales with the Lipschitz constant of the neural network, and upper bounds for estimating this constant can grow with the size of the network \cite{virmaux2018lipschitz,fazlyab2019efficient}. However, as our adapters are small 2-layer MLPs, estimates of this constant suggest we can obtain better generalization with fewer training samples~\cite{jordan2020exactly, yoshida2017spectral, neyshabur2017pac,gouk2021regularisation}. In Section~\ref{sec:accessible_results}, we later show this corresponds to better data efficiency.

\vspace{-0.25cm}
\section{Experiments}
\vspace{-0.225cm}
We now validate that \name{} enables effective and efficient group robustness. First, in Section~\ref{sec:main_results}, we evaluate the effectiveness of \name{} against efficient methods to improve foundation model inference without finetuning. We study whether the approach consistently improves worst-group accuracy and group robustness over zero-shot classification, how \name{} compares against additional efficient methods that only require foundation model embeddings, and whether \name{} scales to a variety of foundation model architectures. Next, in Section~\ref{sec:accessible_results}, we study the efficiency of \name{} against effective state-of-the-art group robustness approaches. 
For popular group robustness benchmarks, we find that \name{} not only achieves state-of-the-art (SoTA) robustness over other methods in the pretrained foundation model regime, but also comparable to SoTA  robustness over standard models used in these benchmarks. However, by only training lightweight adapters, we enable such performance with fewer training parameters and higher data efficiency.


\vspace{-0.125cm}
\subsection{Robustness comparison for efficient foundation model methods}
\label{sec:main_results}
\vspace{-0.125cm}

To first judge the effectiveness of \name{}, we evaluate the method across the same set of initial robustness benchmarks and foundation model architectures discussed in Section~\ref{sec:problem}. As in prior group robustness evaluation, we do not assume training groups labels, but do assume group labels in validation data for hyperparameter tuning and model selection \cite{koh2021wilds}. We include experimental details for all models and hyperparameters in Appendix~\ref{appendix:experimental_details}.

As baselines, we compare against zero-shot classification \cite{radford2021learning}, ERM linear probing \cite{radford2021learning, kumar2022fine}, and ERM adapter training \cite{gao2021clip}. We also compare against recent methods designed to improve downstream transfer in related settings, while similarly only requiring pretrained model embeddings: 
\begin{itemize}[leftmargin=*]
    \item \textbf{Weight space ensembling (WiSE-FT)} \cite{wortsman2021robust}, which first trains a linear classifier with standard ERM, and then ensembles the classifier outputs with the initial zero-shot predictions. While proposed for both training linear classifiers and finetuning the original weights of a foundation model, we focus on the linear classifier version for fair comparison in our setting. 
    \item \textbf{Deep feature reweighting (DFR)} \cite{kirichenko2022last}, which first trains a linear probe on embeddings computed from a pretrained model over group-balanced data. As we do not assume training group labels, we first infer groups using zero-shot classification with foundation model embeddings. As in prior work \cite{liu2021just, zhang2022correct}, we treat the incorrect and correctly classified samples as proxies for different groups. 
\end{itemize}

Finally, assuming we have validation group labels, we know what groups could plausibly be in our test data. We thus also compare against \textbf{group-informed prompting (Group Prompt ZS)}, which performs zero-shot classification using prompts with group information (\eg{} ``a waterbird on a land background'').

\begin{table}[!t]
\caption{\small Evaluation of methods for improving group robustness of CLIP models. Across representative benchmarks and CLIP models, contrastive adapters consistently improve worst-group accuracy over zero-shot classification (by 10.2 to 76.0 pp).  \textbf{1st} / \uline{2nd} best worst-group (WG) and robustness gaps \textbf{bolded} / \uline{underlined}; mean over three seeds.}
\label{table:main_results}
\centering
\begin{adjustbox}{width=\textwidth}
\begin{tabular}{@{}llccbccbccbccb@{}}
\toprule
                               &                     & \multicolumn{3}{c}{Waterbirds}          & \multicolumn{3}{c}{CelebA}             & \multicolumn{3}{c}{BREEDS Living-17}    & \multicolumn{3}{c}{CIFAR-10.02}               \\ \cmidrule(l){3-5} \cmidrule(l){6-8}
                               \cmidrule(l){9-11}
                               \cmidrule(l){12-14}
                         & Method / Acc. (\%)    & WG   & Avg & Gap           & WG   & Avg & Gap          & WG   & Avg & Gap           & WG   & Avg      & Gap           \\ \midrule
\parbox[t]{2mm}{\multirow{8}{*}{\rotatebox[origin=c]{90}{CLIP ResNet-50}}}
& Zero-shot (ZS)      & 36.6          & 92.2    & 55.6          & 74.0            & 81.9    & 7.9          & 6.0           & 86.7    & 80.7          & 39.1          & 69.9          & 30.8          \\
                               & Group Prompt ZS     & 55.9          & 87.8    & 31.9          & 70.8          & 82.6    & 11.8         & 30.0          & 90.6    & 60.6          & N/A           & N/A           & N/A           \\
                               & ERM Linear Probe    & 7.9           & 93.5    & 85.6          & 11.9          & 94.7    & 82.8         & 53.3          & 90.8    & 37.5          & 51.3          & 77.7          & 26.4          \\
                               & ERM Adapter         & 60.8          & 96.0      & 35.2          & 36.1          & 94.2    & 58.1         & \textbf{70.7} & 94.0      & \textbf{23.3} & \textbf{68.8} & 86.0          & \textbf{17.2} \\
                               & WiSE-FT             & 49.8          & 91.0      & 41.2          & 85.6          & 88.6    & 3.0          & 53.3          & 90.8    & 37.5          & 58.2          & 79.1          & 20.9          \\
                               & DFR (Subsample)     & {\ul 63.9}    & 91.8    & {\ul 27.9}    & 76.9          & 92.5    & 15.6         & 46.7          & 89.4    & 42.7          & 45.0          & 75.0          & 30.0          \\
                               & DFR (Upsample)      & 51.3          & 92.4    & 41.1          & {\ul 89.6}    & 91.8    & {\ul 2.2}    & 44.0          & 86.4    & 42.4          & 38.5          & 77.9          & 39.4          \\
                               & \textbf{Contrastive Adapter} & \textbf{83.7} & 89.4    & \textbf{5.7}  & \textbf{90.0} & 90.7    & \textbf{0.7} & {\ul 62.0}    & 90.9    & {\ul 28.9}    & {\ul 60.7}    & 80.9          & {\ul 20.2}    \\ \midrule
\parbox[t]{2mm}{\multirow{8}{*}{\rotatebox[origin=c]{90}{CLIP ViT-L/14}}}
& Zero-shot (ZS)      & 25.7          & 87.3    & 61.6          & 62.1          & 71.9    & 9.8          & 4.0           & 86.6    & 82.6          & 72.0          & 93.2          & 21.2          \\
                               & Group Prompt ZS     & 27.4          & 85.5    & 58.1          & 72.4          & 81.8    & 9.4          & 48.0          & 96.6    & 48.6          & N/A           & N/A           & N/A           \\
                               & ERM Linear Probe    & 65.9          & 97.6    & 31.7          & 28.3          & 94.7    & 66.4         & \textbf{84.0} & 98.6    & {\ul 14.6}    & \textbf{87.5} & 96.1          & \textbf{8.6}  \\
                               & ERM Adapter         & {\ul 78.4}    & 97.8    & 19.4          & 36.7          & 94.2    & 57.5         & {\ul 82.7}    & 98.2    & 15.5          & {\ul 87.0}    & 96.9          & 9.9           \\
                               & WiSE-FT             & 65.9          & 97.6    & 31.7          & {\ul 80.0}      & 87.4    & {\ul 7.4}    & \textbf{84.0} & 98.6    & {\ul 14.6}    & \textbf{87.5} & \textbf{97.0} & {\ul 9.5}     \\
                               & DFR (Subsample)     & 51.9          & 95.7    & 43.8          & 76.3          & 92.1    & 15.8         & \textbf{84.0} & 98.5    & \textbf{14.5} & 85.5          & 96.6          & 11.1          \\
                               & DFR (Upsample)      & 65.9          & 96.1    & {\ul 30.2}    & 83.7          & 91.2    & 7.5          & 78.7          & 97.3    & 18.6          & 72.5          & 93.8          & 21.3          \\
                               & \textbf{Contrastive Adapter} & \textbf{86.9} & 96.2    & \textbf{9.3} & \textbf{84.6} & 90.4    & \textbf{5.8} & 80.0          & 97.5    & 17.5          & 82.2          & 96.1          & 13.9          \\ 
                               \bottomrule
\end{tabular}
\end{adjustbox}
\vspace{-0.125cm}
\end{table}


\begin{wrapfigure}{r}{0.45\textwidth} 
    \centering
    \includegraphics[width=0.425\textwidth]{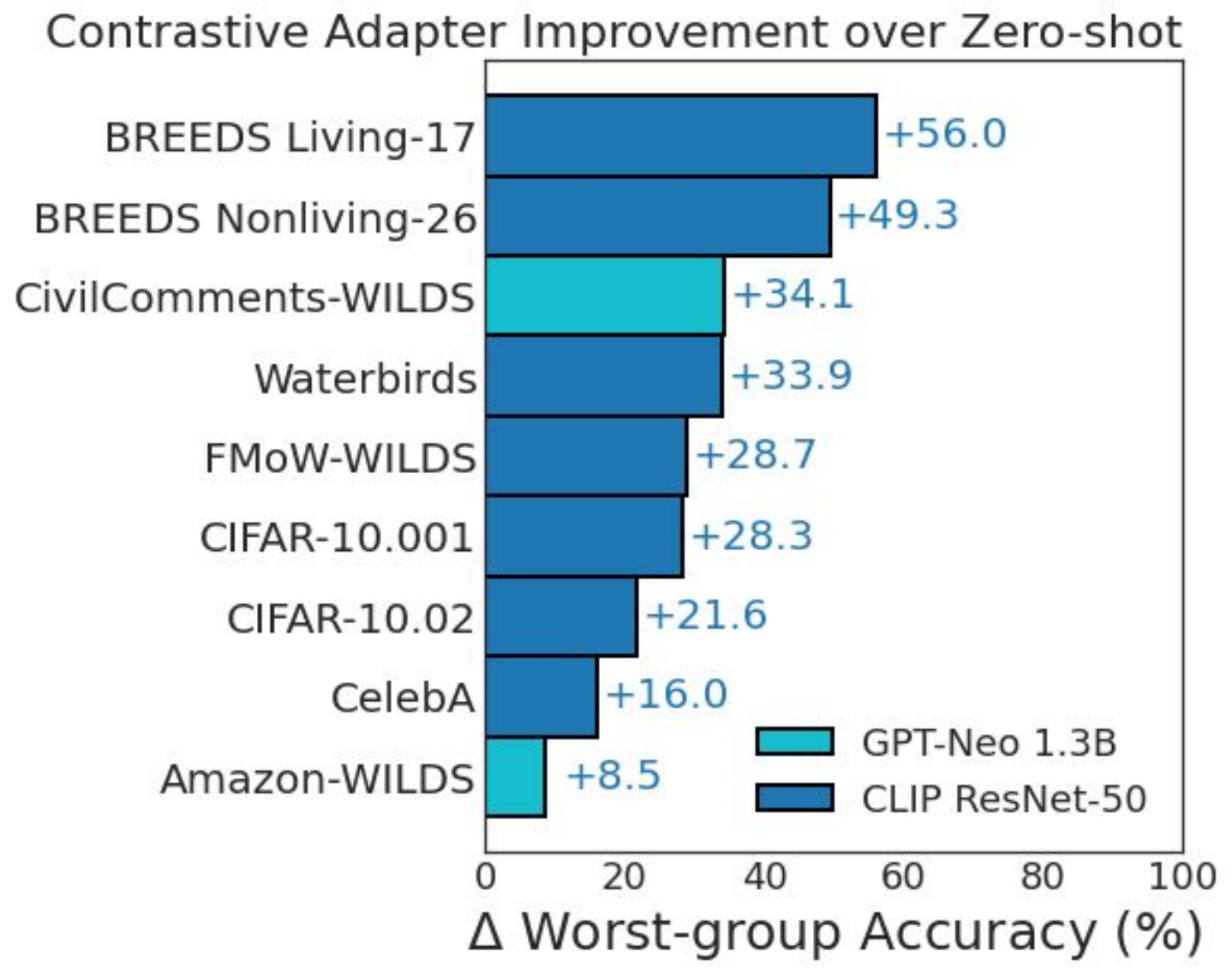}
    \vspace{-0.25cm}
    \caption{Across 9 group robustness benchmarks, \name{} consistently improves worst-group accuracy over foundation model zero-shot classification.}
    \vspace{-0.625cm}
    \label{fig:overall_improvement}
\end{wrapfigure}

\header{Consistent robustness improvements over zero-shot} In Figure~\ref{fig:overall_improvement} we report \name{}'s relative gains in worst-group accuracy over zero-shot classification on all 9 robustness benchmarks. Unlike prior adapter training approaches, \name{} consistently improves group robustness over zero-shot classification, achieving 8.5 to 56.0 pp higher worst-group accuracy.

\header{Representative dataset evaluation} In Table~\ref{table:main_results} we compare \name{} against other lightweight methods for improving robustness.  
We evaluate across group-imbalanced and balanced training data with spurious confounder, subclass, and data source group shifts, using CLIP ResNet-50 (RN-50) and CLIP ViT-L/14 embeddings. 
On average, contrastive adapters improve worst-group accuracy by 12.4 and 4.1 pp over the next best methods for CLIP RN-50 and ViT-L/14 respectively.

\begin{table}[!t]
\vspace{-0.25cm}
\tabcolsep=0.1cm
\caption{\small On the Waterbirds dataset, contrastive adapters consistently improve group robustness across various vision-language foundation models (CLIP \cite{radford2021learning}, CLOOB \cite{furst2021cloob}) and backbones (ResNets and ViTs).}
\label{table:transfer_across_clip}
\centering
\begin{adjustbox}{width=\textwidth}
\begin{tabular}{@{}lccbccbccbccbccb@{}}
\toprule
                    & \multicolumn{3}{c}{CLOOB RN-50}               & \multicolumn{3}{c}{CLOOB RN-50x4}             & \multicolumn{3}{c}{CLIP RN-101}               & \multicolumn{3}{c}{CLIP ViT-B/32}              & \multicolumn{3}{c}{CLIP ViT-B/16}              \\  \cmidrule(l){2-4} 
                    \cmidrule(l){5-7}
                    \cmidrule(l){8-10}
                    \cmidrule(l){11-13}
                    \cmidrule(l){14-16}
Accuracy (\%)       & WG            & Avg           & Gap          & WG            & Avg           & Gap          & WG            & Avg           & Gap          & WG            & Avg           & Gap          & WG            & Avg           & Gap          \\ \midrule
Zero-shot           & 41.6          & 60.4          & 18.8         & 24.1          & 51.1          & 27           & 33.6          & \textbf{90.0} & 56.4         & 47.0          & \textbf{88.8} & 41.8         & 34.0          & 88.1          & 54.1         \\
Contrastive Adapter & \textbf{83.0} & \textbf{86.8} & \textbf{3.8} & \textbf{85.8} & \textbf{88.5} & \textbf{2.7} & \textbf{82.0} & 86.0          & \textbf{4.0} & \textbf{80.7} & 84.2          & \textbf{3.5} & \textbf{83.1} & \textbf{90.9} & \textbf{7.8} \\ \bottomrule
\end{tabular}
\end{adjustbox}
\vspace{-0.5cm}
\end{table}

\header{Transfer across architectures}
To further evaluate \name{}'s effectiveness, we study how the prior improvements transfer to different foundation models. In Table~\ref{table:transfer_across_clip}, we find that contrastive adapters substantially improve group robustness for other foundation models such as CLOOB~\cite{furst2021cloob}. \Name{} also scales across foundation model backbones, raising worst-group accuracy by 33.7 to 61.7 pp while only training models with 0.52\% to 1.03\% of the original FM parameter count. 











\vspace{-0.25cm}
\subsection{Measuring efficiency among effective group robustness solutions}
\label{sec:accessible_results}

\begin{table}[!t]
\tabcolsep=0.1cm
\caption{\small On popular Waterbirds and CelebA benchmarks, contrastive adapters achieve near state-of-the-art worst-group accuracy (WG Acc.) with $\leq$1\% of the trainable parameters. $\Delta$Acc. is percentage point gap with prior SoTA. \textbf{1st} / \uline{2nd} best metrics \textbf{bolded} / \uline{underlined}. We report numbers from original works, means over three seeds for ours. }
\label{table:sota_comparison}
\centering
\begin{adjustbox}{width=\textwidth}
\begin{tabular}{@{}lccccccc@{}}
\toprule
                           & \multicolumn{1}{l}{}      & \multicolumn{1}{l}{} &              & \multicolumn{2}{c}{Waterbirds}                   & \multicolumn{2}{c}{CelebA}                 \\ \cmidrule(l){5-6}\cmidrule(l){7-8} 
Model                      & \# Trained Params         & \% Params            & Method       & WG Acc. (\%)        & $\Delta$Acc. & WG Acc. (\%)  & $\Delta$Acc. \\ \midrule
\multirow{7}{*}{ResNet-50} & \multirow{7}{*}{25557032} & \multirow{7}{*}{100} & EIIL~\cite{creager2021environment}         & 78.7                & -10.3                      & 83.3          & -6.5                       \\
                           &                           &                      & CIM~\cite{taghanaki2021robust}          & 83.6                & -5.4                       & 83.6          & -6.2                       \\
                           &                           &                      & JTT~\cite{liu2021just}         & 86.7                & -2.3                       & 81.1          & -8.7                       \\
                           &                           &                      & SUBY~\cite{idrissi2021simple}         & 82.4                & -6.6                       & 79.9          & -9.9                       \\
                           &                           &                      & RWY~\cite{idrissi2021simple}          & 86.1                & -2.9                       & 82.9          & -6.9                       \\
                           &                           &                      & CNC~\cite{zhang2022correct}          & {\ul 88.5}                & -0.5                       & 88.8          & -1.0                       \\
                           &                           &                      & \textbf{SSA}~\cite{nam2022spread} & \textbf{89.0} & 0.0                        & {\ul 89.8}    & 0.0                        \\ \midrule
Adapter + CLIP RN-50       & 263424                    & 1.03                 & Ours         & 83.7                & -5.3                       & \textbf{90.0} & 0.2                        \\
Adapter + CLIP ViT-L/14    & 197632                    & 0.77                 & Ours         & 86.9 & -2.1                       & 84.6          & -5.2                       \\ \bottomrule
\end{tabular}
\end{adjustbox}
\end{table}

In the previous section, we showed how contrastive adapters could significantly improve group robustness for foundation models. We now expand on \name{}'s efficiency. We find that for group robust classification in general, \name{} can achieve state-of-the-art performance despite only training $\leq$1\% of the usual model parameters. The lightweight nature of \name{} also leads to better data efficiency than existing state-of-the-art approaches.

\header{Robustness comparison to state-of-the-art methods} In Table~\ref{table:sota_comparison}, we evaluate how \name{} with CLIP RN-50 and ViT-L/14 embeddings compares to current state-of-the-art robustness techniques. For evaluation, we use the popular Waterbirds and CelebA datasets. The existing methods train an ImageNet-pretrained ResNet-50. 
We find that on both datasets, \name{} achieves comparable worst-group accuracy to recent group robustness methods, despite only training $\leq$1\% of their parameters. Notably, \name{} outperforms some of these approaches by up to 5.0 and 10.1 pp for Waterbirds and CelebA respectively. Contrastive adapting also falls short of the state-of-the-art Spread Spurious Attribute (SSA) method by 2.1 pp on Waterbirds, but achieves 0.2 higher worst-group accuracy on CelebA. This suggests contrastive adapters not only effectively improve group robustness for foundation models, but also enable competitive robust classification in general with a fraction of prior approaches' trainable parameters.

\begin{wrapfigure}{r}{0.45\textwidth} 
\vspace{-0.5cm}
    \centering
    \includegraphics[width=0.425\textwidth]{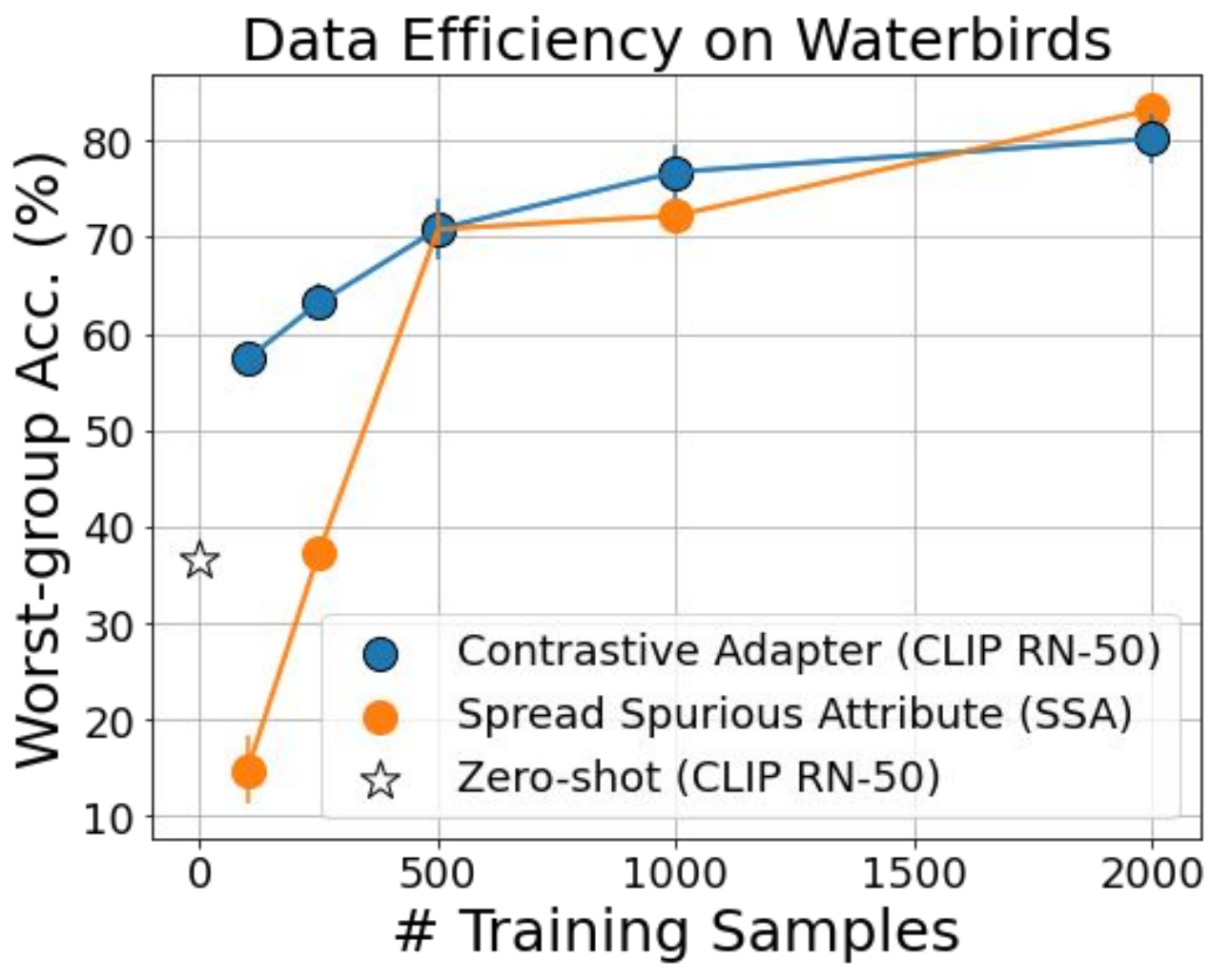}
    \caption{In fewer data regimes, contrastive adapters maintain higher group robustness than zero-shot classification, and significantly outperform standard models trained with the state-of-the-art SSA method.} 
    \vspace{-0.5cm}
    \label{fig:data_efficiency}
\end{wrapfigure}

\header{Data efficiency evaluation} Beyond model parameter count, we also study if the lightweight nature of \name{} transfers to better data efficiency. We compare \name{} to the best performing SSA~\cite{nam2022spread} on subsampled versions of Waterbirds. To evaluate how well methods maintain group robustness with less training data available, we keep group ratios preserved (\ie, for Waterbirds, within each class 5\% of the training samples for that class belong to a minority group~\cite{sagawa2019distributionally}). 


In Figure~\ref{fig:data_efficiency}, we show that \name{} substantially outperforms SSA in lower data regimes. With only 100 and 250 training samples, \name{} outperforms SSA-trained ResNet-50s by 42.6 and 25.9 pp in worst-group accuracy. With 100 training samples (where the smallest group only has 1 sample), \name{} still improves worst-group accuracy by 20.8 pp over zero-shot classification. In contrast, SSA's performance drops significantly, resulting in 17.6 pp lower worst-group accuracy than zero-shot classification. To connect this greater robustness in lower data regimes to our discussion of robust adapter generalization in Section~\ref{sec:method_algorithm}, we also empirically estimate the Lipschitz constant of the trained adapter. We use prior methods for neural network lipschitz constant estimation~\cite{fazlyab2019efficient}. We find the constant to be 29.3, which is much lower than the constant reported for networks with much larger architectures (\eg~as applicable to the ResNet-50 architecture used in state-of-the-art group robustness methods such as SSA) \cite{fazlyab2019efficient, virmaux2018lipschitz}.

\section{Discussion: limitations and societal impact}

\header{Limitations} While in this work, we demonstrated that we can substantially improve the group robustness of foundation model classification without any finetuning of the original model, several limitations still exist. First, this does not imply that we can get desirable performance in general without additional retraining. To obtain high worst-group performance in general, we are upper-bounded by whether the pretrained embeddings do contain the information needed to classify all groups. While our study suggests that in many cases they do carry this information---which can surprising given that the zero-shot classification with the same embeddings results in poor group robustness---in other situations the pretrained embeddings may lack this information. For example, if downstream task data is very different in distribution from the pretraining data, then the pretrained foundation model embeddings may not be sufficient to work with. While more efficient ways to improve robustness can democratize foundation model use, further finetuning may still be needed.

We also emphasize that our approach is a simple first-step method to improving the group robustness over existing baseline approaches. This is motivated by our observation that foundation model zero-shot classification may not be group robust, and that we would like to both (i) improve performance of these models when we realize they fail in certain aspects, and (ii) do so efficiently, such that fixing their failures is not bound by who can conduct costly retraining procedures. While contrastive adapting critically \emph{consistently} improves worst-group performance over zero-shot classification, it does not always outperform all other approaches. We found this to depend on the training data characteristics; contrastive adapters achieved better results on training data with spurious correlations, and other approaches such as training ERM adapters worked better with balanced groups. We are excited for future work and expect further improvements as this thread of how to efficiently improve FM performance with limited access (\eg{} only pretrained embeddings) is further explored over various training data and deployment settings.  

\header{Societal impact} Finally, we note that it is important to carefully study the learned biases of foundation models, and to devise appropriate solutions evaluated outside of just computational metrics. Due to their promise of widespread and effective downstream transfer, foundation models may have a particularly strong impact on various parts of society. Individuals may get the sense that they can successfully apply these pretrained models to their desired downstream tasks ``out-of-the-box''. However, doing so also risks applying any learned biases of the model. Our work raises this issue with respect to group robustness as motivation for our problem setting, and also notes that additional evaluation beyond average accuracy can shed light on the negative qualities of existing models (where zero-shot FM classification may perform very well on average, but very poorly on certain groups, c.f. Figure~\ref{fig:zeroshot_poor_robustness_all}). However we recognize the limitations of purely computational solutions to addressing group performance disparities in society. Further discussion is needed to describe desirable objectives. We need to better understand foundation models within this context and their potential uses in broader socio-technical systems \cite{bommasani2021opportunities}.



\section{Conclusion}
\label{sec:conclusion}
We study the group robustness of popular foundation models. We find their zero-shot classification may not be robust to various group shifts, 
establish that baseline linear probe and adapter strategies do not reliably improve robustness, and propose a simple adapter strategy to significantly and consistently improve FM robustness without finetuning.
%
This suggests FM embeddings do contain group-relevant information, and we show that we can use FM embeddings to efficiently achieve state-of-the-art robust classification. 
%


%




\newpage

\vspace{-0.1cm}
\section*{Acknowledgments}
\vspace{-0.1cm}
We thank Simran Arora, Megan Leszczynski, Kush Bhatia, Maya Varma, Gautam Machiraju, and Laurel Orr for helpful discussions and feedback.

We gratefully acknowledge the support of NIH under No. U54EB020405 (Mobilize), NSF under Nos. CCF1763315 (Beyond Sparsity), CCF1563078 (Volume to Velocity), and 1937301 (RTML); ARL under No. W911NF-21-2-0251 (Interactive Human-AI Teaming); ONR under No. N000141712266 (Unifying Weak Supervision); ONR N00014-20-1-2480: Understanding and Applying Non-Euclidean Geometry in Machine Learning; N000142012275 (NEPTUNE); Apple, NXP, Xilinx, LETI-CEA, Intel, IBM, Microsoft, NEC, Toshiba, TSMC, ARM, Hitachi, BASF, Accenture, Ericsson, Qualcomm, Analog Devices, Google Cloud, Salesforce, Total, the HAI-GCP Cloud Credits for Research program, the Stanford Data Science Initiative (SDSI),
and members of the Stanford DAWN project: Facebook, Google, and VMWare. The U.S. Government is authorized to reproduce and distribute reprints for Governmental purposes notwithstanding any copyright notation thereon. Any opinions, findings, and conclusions or recommendations expressed in this material are those of the authors and do not necessarily reflect the views, policies, or endorsements, either expressed or implied, of NIH, ONR, or the U.S. Government.

\bibliographystyle{plainnat}
\bibliography{main}

\newpage

\appendix

\section{Expanded zero-shot evaluation for group robustness}
\label{appendix:extended_zeroshot_evaluation}
In this section, we expand on the zero-shot evaluation of various foundation models on group robustness benchmarks discussed in Section~\ref{sec:problem}. We first describe the datasets and models used in Appendix~\ref{appendix:extended_zeroshot_evaluation_additional_details}. We then include results in Appendix~\ref{appendix:extended_zeroshot_evaluation_results}. We find consistent trends of poor group robustness with zero-shot classification, marked by poor worst-group accuracy and large gaps between average and worst-group accuracy.

\subsection{Additional details on robustness datasets and foundation models}
\label{appendix:extended_zeroshot_evaluation_additional_details}

\header{Datasets} To benchmark zero-shot group robustness, we use a diverse set of datasets with group shifts from prior robustness literature. We describe them below and include details on size of groups and type of group shift in Table~\ref{table:dataset_details}:
\begin{itemize}[leftmargin=*]
    \item \textbf{Waterbirds}~\cite{WelinderEtal2010, sagawa2019distributionally}. We classify images by bird type. Each class $\in \{ \texttt{waterbird}, \texttt{landbird} \}$ carries two groups: birds on water backgrounds, and birds on land backgrounds.
    \item \textbf{CelebA}~\cite{liu2015deep, sagawa2019distributionally}. We classify images by celebrity hair color. Each class $\in \{ \texttt{not blond}, \texttt{blond}\}$ carries two groups: celebrities labeled as male, and celebrities labeled as female.
    \item \textbf{BREEDS (\texttt{Living-17}, \texttt{Nonliving-26})} ~\cite{santurkar2020breeds}. For the \texttt{Living-17} and \texttt{Nonliving-26} datasets in the BREEDS benchmark sourced from ImageNet~\cite{santurkar2020breeds}, we classify images by one of several categories. Each class is a coarse category consisting of multiple fine-grained groups. Groups in the same class may be visually distinct (\eg{} the \texttt{ape} class includes images of gibbons and gorillas).
    While the original benchmark evaluates how classifiers trained on seen \texttt{source} groups generalize to unseen \texttt{target} groups, we adapt the datasets for our group robustness setting by adding 5\% of the images in each \texttt{target} group to the \texttt{source} groups, and evaluating worst-group accuracy over all \texttt{source} and \texttt{target} groups.
    \item \textbf{CIFAR-10.001}, \textbf{CIFAR-10.02} \cite{Krizhevsky2009LearningML, recht2018cifar, lu2020harder}. We classify images by one of 10 categories. We combine CIFAR-10~\cite{Krizhevsky2009LearningML} and either CIFAR-10.1~\cite{recht2018cifar} or CIFAR-10.2~\cite{lu2020harder}, which are collected from different sources. The new datasets' classes carry two groups determined by the source dataset.
    \item \textbf{FMoW-WILDS}~\cite{christie2018functional, koh2021wilds}. We classify satellite images into one of 62 building or land-use categories (\eg{} \texttt{airport}, \texttt{zoo}). Each images belongs to one of five groups based on continental region. 
    To test group robustness, we compare the accuracies over all samples in each group as in the WILDS benchmark \cite{koh2021wilds}. 
    We also evaluate only over test images from the same time period as training images (the ``IID'' split in the original WILDS benchmark \cite{koh2021wilds}). 
    \item \textbf{CivilComments-WILDS}~\cite{DBLP:journals/corr/abs-1903-04561, koh2021wilds}. We classify if a text comment is toxic or not. Samples are organized into 8 groups based on mention of a demographic identity (\eg{} ``female'', ``LGBTQ'').

    \item \textbf{Amazon-WILDS}~\cite{ni-etal-2019-justifying, koh2021wilds}. We classify if an online text review is positive or negative. Reviews are organized into different groups based on the product category (\eg{} \texttt{books}, \texttt{electronics}). We adapt this dataset from the official Amazon-WILDS split by using the \texttt{category\_subpopulation} split. We also map the original class labels, which are star-ratings from 1 to 5, to positive or negative reviews by discarding samples with a 3-star rating, and re-labeling 1- and 2-star ratings as \texttt{negative} and 4- and 5-start ratings as \texttt{positive}.
    
\end{itemize}


\header{Foundation models}
For image datasets, we evaluate pretrained CLIP~\cite{radford2021learning} and CLOOB~\cite{furst2021cloob} vision-language models using publicly available weights\footnote{CLIP: \href{https://github.com/openai/CLIP/blob/main/clip/clip.py}{https://github.com/openai/CLIP/blob/main/clip/clip.py}}\footnote{CLOOB: \href{https://ml.jku.at/research/CLOOB/downloads/checkpoints/}{https://ml.jku.at/research/CLOOB/downloads/checkpoints/}}. We evaluate 7 available CLIP models: 3 ResNet image encoder backbones (\texttt{RN-50}, \texttt{RN-101}, \texttt{RN-50x4}), and 5 Vision Transformer image encoder backbones: (\texttt{ViT-B/32}, \texttt{ViT-B/16}, \texttt{ViT-L/14}, \texttt{ViT-L/14@336px}) and 2 CLOOB models (all available: \texttt{RN-50}, \texttt{RN-50x4}). For text datasets, we evaluate 2 pretrained GPT-Neo~\cite{gpt-neo} text models trained on the Pile~\cite{gao2020pile} (\texttt{GPT-Neo-125M}, \texttt{GPT-Neo-1.3B}) available on HuggingFace\footnote{GPT-Neo 125M: \href{https://huggingface.co/EleutherAI/gpt-neo-125M}{https://huggingface.co/EleutherAI/gpt-neo-125M}}\footnote{GPT-Neo 1.3B: \href{https://huggingface.co/EleutherAI/gpt-neo-1.3B}{https://huggingface.co/EleutherAI/gpt-neo-1.3B}}.

\subsection{Group robustness results}
\label{appendix:extended_zeroshot_evaluation_results}

\begin{figure}[!h]
  \centering
  \includegraphics[width=1\textwidth]{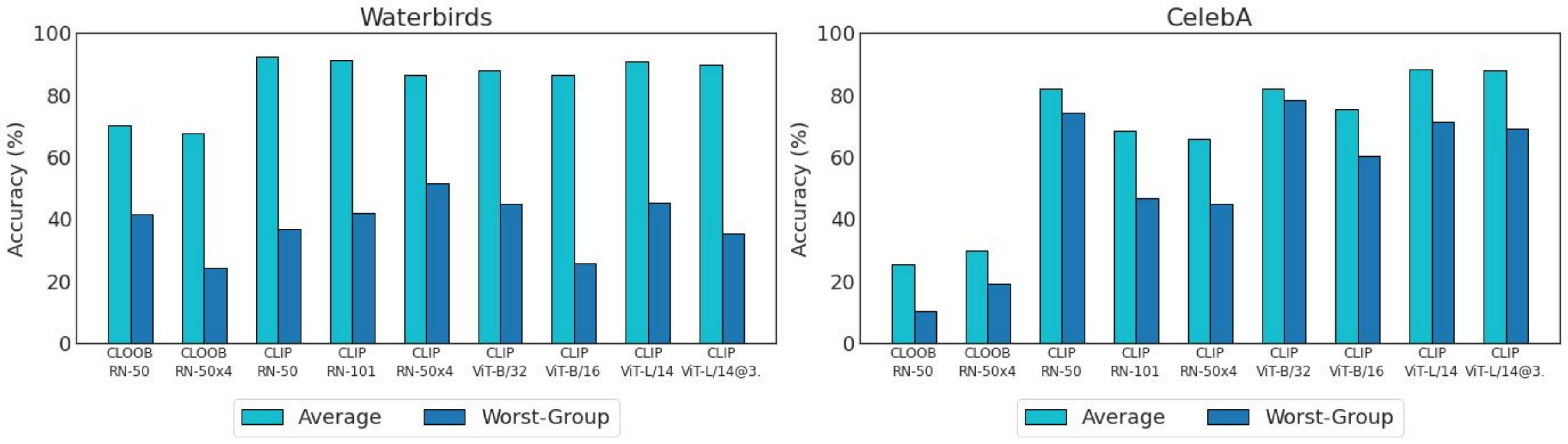}

  \includegraphics[width=1\textwidth]{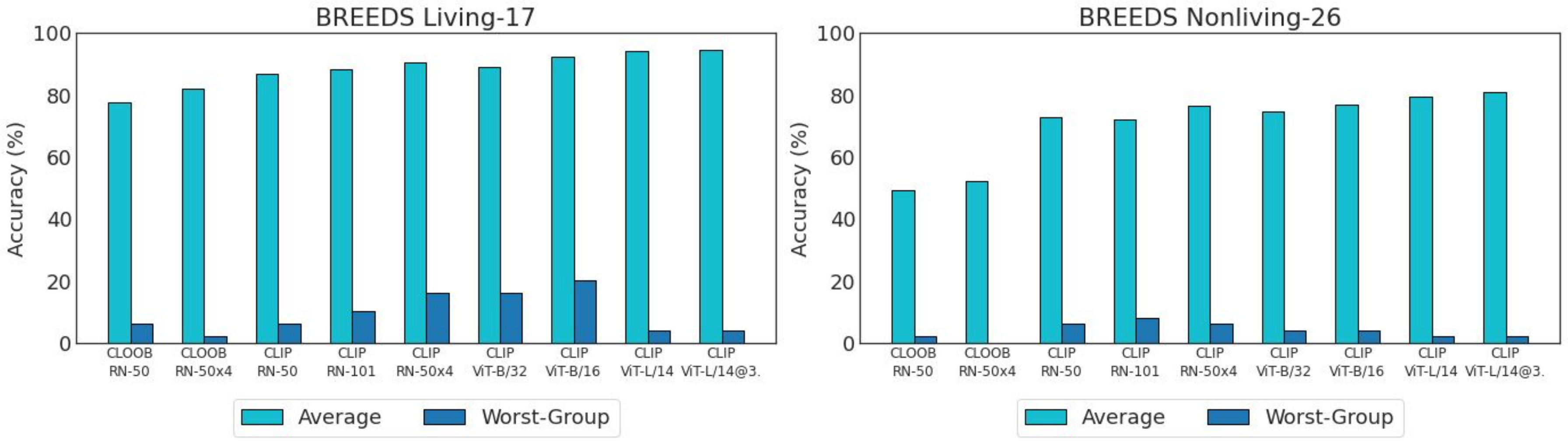}

  \includegraphics[width=1\textwidth]{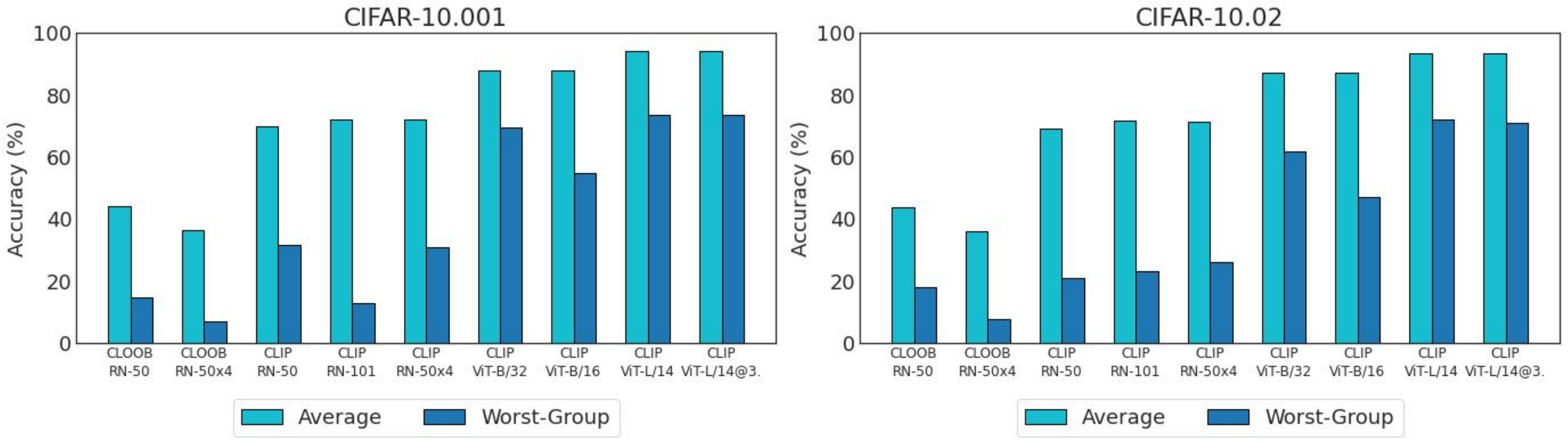}

  \includegraphics[width=1\textwidth]{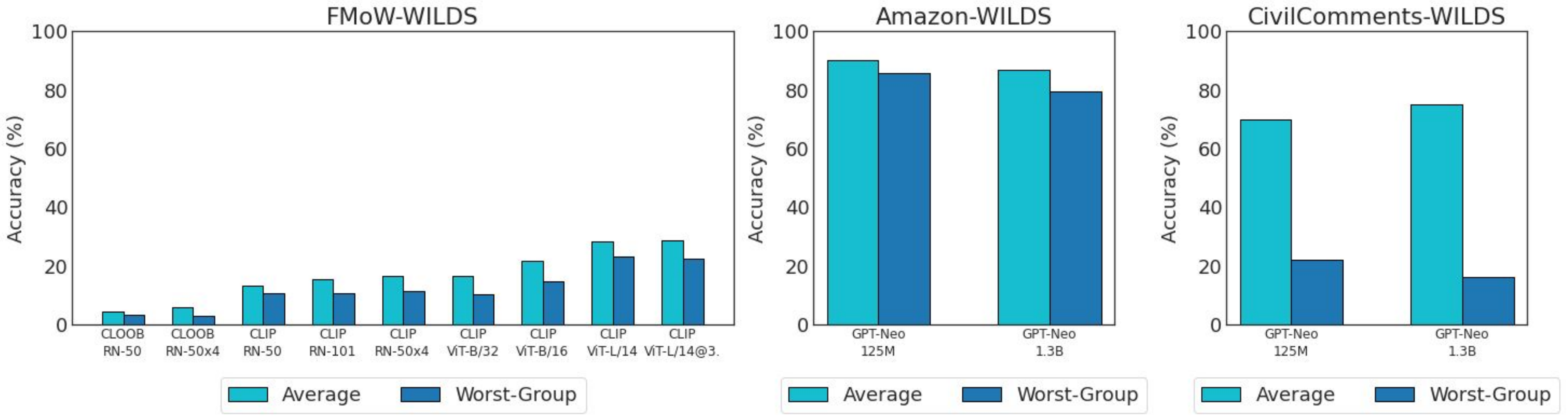}
  \caption{Foundation model zero-shot classification accuracies. We find poor zero-shot group robustness across datasets and models via large gaps between average and worst-group accuracies.}
  \vspace{-0.25cm}
  \label{fig:zeroshot_poor_robustness_all}
\end{figure}

In Figure~\ref{fig:zeroshot_poor_robustness_all}, we chart worst-group and average accuracies achieved by various zero-shot foundation models across the group robustness datasets. Larger gaps between accuracies, \ie{} high average accuracy yet low worst-group accuracy, indicate poor group robustness. In aggregate, on all datasets except FMoW-WILDS and Amazon-WILDS, we observe a shared pattern of noticeable gaps between average and worst-group accuracy, suggesting that zero-shot classification with popular foundation models may not be group robust. We perform zero-shot classification as described in Section~\ref{sec:problem}. As recommended by \citet{radford2021learning}, for each dataset we consider several prompt templates. We engineer prompts by using the single best template based on validation worst-group accuracy. In Appendix~\ref{appendix:experimental_details_prompts} we include a full list of prompts used.

\begin{table}[!t]
\caption{Group robustness datasets, source of group shift, and group sizes.}
\label{table:dataset_details}
\centering
\begin{adjustbox}{width=0.8\textwidth}
{\small
\begin{tabular}{@{}llccc@{}}
\toprule
                    &             & \multicolumn{3}{c}{(Class-wise) Group Size} \\ \cmidrule(l){3-5} 
Dataset             & Group Shift & Largest     & Smallest     & Class-Wise?    \\ \midrule
Waterbirds          & Confounder  & 1057        & 56           & Yes            \\
CelebA              & Confounder  & 22880       & 1387         & Yes            \\
BREEDS Living-17    & Subclass    & 1076        & 1009         & Yes            \\
BREEDS Nonliving-26 & Subclass    & 1043        & 712          & Yes            \\
CIFAR-10.001        & Data source & 1000        & 114          & Yes            \\
CIFAR-10.02         & Data source & 4039        & 431          & Yes            \\
FMoW-WILDS          & Subclass    & 34816       & 1582         & No             \\
Amazon-WILDS        & Subclass    & 496127      & 110          & No             \\
CivilComments-WILDS & Confounder  & 4962        & 1003         & Yes            \\ \bottomrule
\end{tabular}
}
\end{adjustbox}
\end{table}

\section{Contrastive adapter implementation details}
\label{appendix:extended_related_work}

We provide further details on the adapter architecture and training sampling.

\begin{wrapfigure}{r}{0.3\textwidth} 
\vspace{-1cm}
    \centering
    \includegraphics[width=0.25\textwidth]{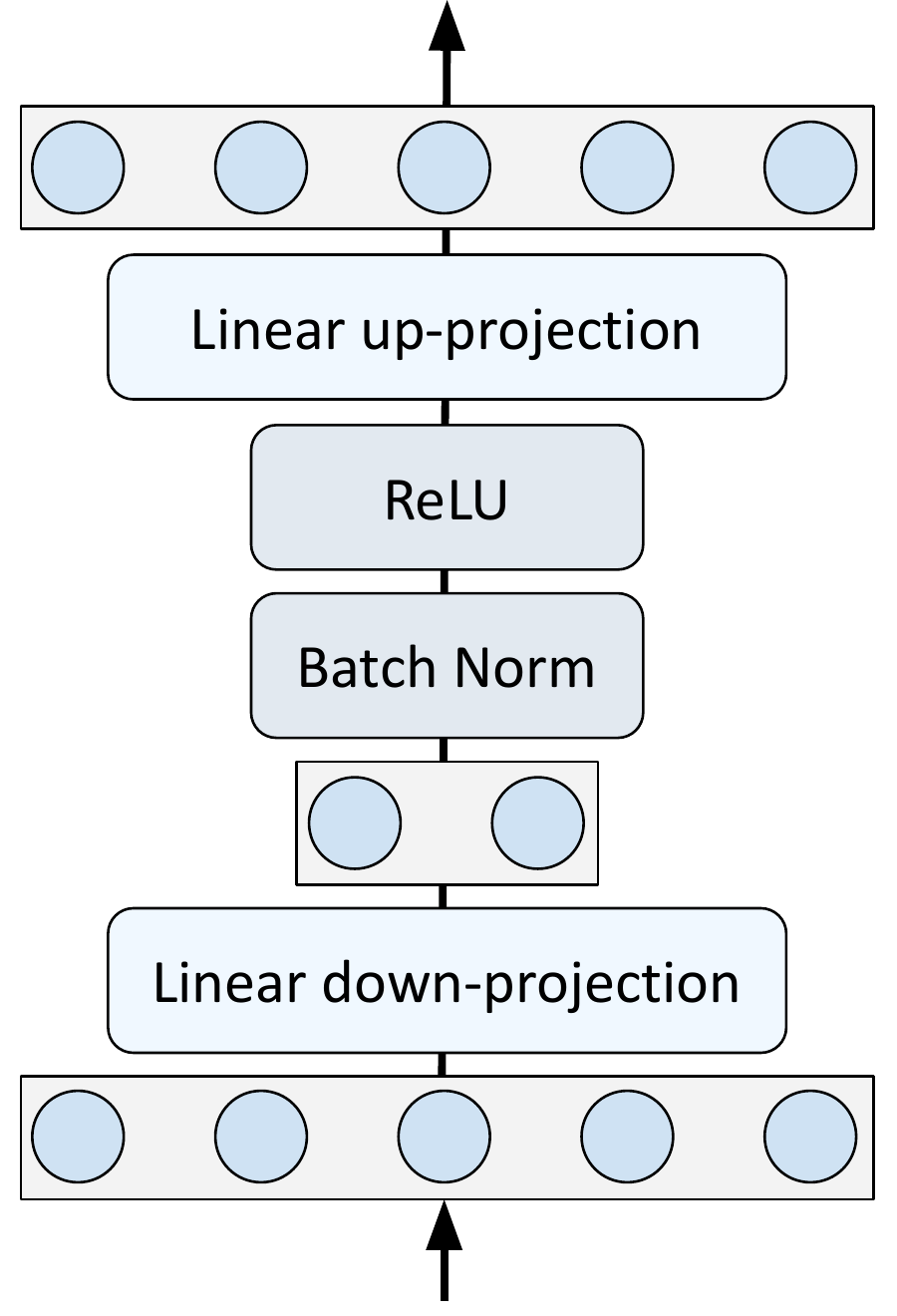}
    \vspace{-0.25cm}
    \caption{Adapter architecture} 
    \vspace{-2.0cm}
    \label{fig:adapter_arch}
\end{wrapfigure}

\subsection{Adapter architecture}
Similar to prior works~\cite{pmlr-v97-houlsby19a, gao2021clip}, the adapters we use are bottleneck 2-layer multilayer perceptrons (MLPs). We set the input dimension and output dimensions as the same as the pretrained foundation model embedding dimension, and pick a smaller dimension for the hidden layer (frequently 128, although this was chosen as a heuristic and not tuned). We also experimented with using a single residual connection~\cite{he2016deep} and batch normalization layer~\cite{ioffe2015batch} between the input and output layers, but only found the latter to be helpful. Pytorch-like pseudocode is given below. The adapter is visualized in Figure~\ref{fig:adapter_arch}.

\begin{lstlisting}[language=Python]
import torch.nn as nn

class Adapter(nn.Module):
    def __init__(self, input_dim, hidden_dim):
        super().__init__()
        self.arch = nn.Sequential(
            nn.Linear(input_dim, hidden_dim),
            nn.BatchNorm1d(hidden_dim),
            nn.ReLU(),
            nn.Linear(hidden_dim, input_dim)
        )
    def __forward__(self, x):
        return self.arch(x)
\end{lstlisting}

\subsection{Adapter training sampling}
Recall in Section~\ref{sec:method_algorithm} that we train the adapter with both a supervised contrastive loss over specifically sampled ``contrastive batches'', and a cross-entropy loss over resampled batches, both over the fixed pretrained foundation model embeddings. We use the foundation model's zero-shot predictions to guide sampling for both. We outline the algorithms for sampling training batches in Algorithm~\ref{algo:contrastive_batch_sampling} and Algorithm~\ref{algo:resampled_sampling}. We then train an adapter by applying the contrastive loss (Eq.~\ref{eq:contrastive_adapter}) and the cross-entropy loss (Eq.~\ref{eq:cross_entropy_adapter}) over these batches in Algorithm~\ref{algo:contrastive_adapting}.

\begin{algorithm}[H]
\caption{Contrastive batch sampling}
\label{algo:contrastive_batch_sampling}
\begin{algorithmic}[1]
    \Input Training dataset sample embeddings $U = \{u_n\}_{n=1}^N$. Ground-truth class labels $Y = \{y_n\}_{n=1}^N$. Foundation model zero-shot predictions $\hat{Y} = \{\hat{y}_n\}_{n=1}^N$. 
    \Require Number of positives $P$ per anchor. Number of negatives $M$ per anchor. Number of nearest neighbors $M^*$ per anchor to sample negatives from.
    \State{Initialize set of contrastive batches $B = \{\}$}
    \For{anchor $u_a \in \{u_i \in U : \hat{y}_i \neq y_i \}$}
    \Statex{\;\;\;\;\;\textbf{(Positive sampling)}}
    \State {Sample $P$ positives $\{u_p\}_{p=1}^P$ uniform-randomly from $U$ where $\hat{y}_p = y_p$ (and $\hat{y}_p \neq \hat{y}_a)$}
    \Statex{\;\;\;\;\;\textbf{(Negative sampling)}}
    \State {Sample $M$ negatives $\{u_m\}_{m=1}^M$ by computing the $M^*$ sample embeddings with the highest }
    \Statex{\;\;\;\;\;cosine similarity to $u_a$ where $y_m \neq y_a$, then randomly sampling $M$ of these embeddings}
    \State{Update contrastive batch sets $B \leftarrow B \cup \left( u_a,\{u_p\}_{p=1}^P, \{u_m\}_{m=1}^M  \right)$}
    \EndFor
\end{algorithmic}

\end{algorithm}
\vspace{-0.5cm}

\begin{algorithm}[H]
\caption{Resampled training set sampling}
\label{algo:resampled_sampling}
\begin{algorithmic}[1]
    \Input Training dataset sample embeddings $U = \{u_n\}_{n=1}^N$. Ground-truth class labels $Y = \{y_n\}_{n=1}^N$. Foundation model zero-shot predictions $\hat{Y} = \{\hat{y}_n\}_{n=1}^N$. All unique classes $C$.
    \State{Initialize resampled training samples $U^* = \{\}$}
    \For{class $c \in C$}
    \State {Identify incorrect samples $U^- = \{u_i\}$ where $\hat{y}_i \neq c$}
    \State {Identify correct samples $U^+ = \{u_i\}$ where $\hat{y}_i = c$}
    \State {Obtain upsampled samples $\tilde{U}^{-}$ by uniform-randomly sampling from $U^-$ s.t. $|\tilde{U}^{-}| = |U^+|$ }
    \State{Update resampled samples $U^* \leftarrow U^* \cup \big( \tilde{U}^{-} \cup U^+  \big)$}
    \EndFor
\end{algorithmic}
\end{algorithm}
\vspace{-0.5cm}

\begin{algorithm}[H]
\caption{Contrastive adapting}
\label{algo:contrastive_adapting}
\begin{algorithmic}[1]
    \Input Set of contrastive batches $B$, resampled training samples $U^*$, number of epochs $K$. 
    \State{Randomly initialize adapter $f_\theta$}
    \For{epoch $1, \ldots, K$}
    \State {Sample contrastive batch $\{b\}$ from $B$}
    \State {Sample randomly-shuffled minibatch of samples $\{u\}$ from $U^*$}
    \State {Update $f_\theta$ with Equation~\ref{eq:contrastive_adapter} over $\{b\}$}
     \State {Update $f_\theta$ with Equation~\ref{eq:cross_entropy_adapter} over $\{u\}$}
    \EndFor
\end{algorithmic}
\end{algorithm}


\section{Additional experimental details}
\label{appendix:experimental_details}

\subsection{Model selection and hyperparameters}
\label{appendix:experimental_details_hyperparameters}
For each dataset and method, we use the following hyperparameters. As in prior group robustness work~\cite{koh2021wilds}, we select the best model and hyperparameters based on early stopping that achieves highest worst-group validation accuracy. For all methods and datasets, we train both linear probes and adapters with SGD, and sweep over learning rate $\in \{$1e-3, 1e-4, 1e-5$\}$ and weight decay $\in \{$5e-5, 1e-5, 5e-4$\}$. For adapter classification, we used the default temperature used for zero-shot classification in CLIP~\cite{radford2021learning}. We did not tune the contrastive temperature. Unless noted, we ran all numbers.

For zero-shot predictions to set up training samples, we tried two procedures. We consider both the procedure based on nearest neighbors to class embeddings described in Section~\ref{sec:foundation_model_group_robustness_problem}, and a clustering approach where we use K-means clustering on the pretrained sample embeddings, setting $K$ to be the number of classes. As done in prior work~\cite{sohoni2020no, zhang2022correct}, we do the clustering over UMAP~\cite{mcinnes2018umap}-fitted representations, following prior suggestions that doing so is helpful for clustering over neural network representations. We treat the zero-shot prediction method as a hyperparameter, and select based on highest worst-group validation accuracy. We use the same zero-shot prediction method for each method that requires initial predictions (DFR and Contrastive Adapter, c.f. Table~\ref{table:main_results}) and across all foundation model architectures. This amounts to doing the clustering approach for Waterbirds and CelebA, and the default nearest neighbors approach for all other datasets.

We list hyperparameters for linear probes (Table~\ref{table:linear_probe_hparams}), adapters (Table~\ref{table:adapter_hparams}, both ERM and contrastive), and contrastive-specific hyperparameters (Table~\ref{table:contrastive_adapter_hparams}). We discuss method-specific hyperparameters:
\begin{itemize}[leftmargin=*]
    \item \textbf{Contrastive adapting} requires selecting three additional hyperparameters: the number of positives and negatives, and the number of nearest neighbors to sample negatives from. For these we swept over the following combinations of (number positives, number negatives, number neighbors): (2048, 2048, 2146), (2048, 2048, 4096), (512, 512, 1024).
    \item \textbf{Weight-space ensembling} (WiSE-FT): WiSE-FT requires picking a value $\alpha \in [0, 1]$ to compute a weighted combination of the zero-shot classifier parameters and the trained linear probe parameters. 
    We sweep over intervals of size $0.1$, \ie{} $\alpha \in \{0.0, 0.1, 0.2, 0.3, 0.4, 0.5, 0.6, 0.7, 0.8, 0.9, 1.0\}$.
\end{itemize}

\begin{table}[!h]
\caption{Linear probe hyperparameters}
\label{table:linear_probe_hparams}
\centering
\begin{adjustbox}{width=\textwidth}
{
\begin{tabular}{@{}lccccc@{}}
\toprule
Dataset             & Max Epochs & Learning Rate & Weight Decay & Momentum & Batch Size \\ \midrule
Waterbirds          & 100        & 1e-3        & 5e-5      & 0.9      & 128        \\
CelebA              & 50         & 1e-3        & 5e-5      & 0.9      & 128        \\
BREEDS Living-17    & 100        & 1e-3        & 5e-5      & 0.9      & 128        \\
BREEDS Nonliving-26 & 100        & 1e-3        & 5e-5      & 0.9      & 128        \\
CIFAR-10.001        & 100        & 1e-3        & 5e-5      & 0.9      & 128        \\
CIFAR-10.02         & 100        & 1e-3        & 5e-5      & 0.9      & 128        \\
FMoW-WILDS          & 100        & 1e-3        & 5e-5      & 0.9      & 128        \\
Amazon-WILDS        & 100        & 1e-3        & 5e-5      & 0.9      & 16         \\
CivilComments-WILDS & 100        & 1e-3        & 5e-5      & 0.9      & 16         \\ \bottomrule
\end{tabular}
}
\end{adjustbox}
\end{table}

\begin{table}[!h]
\caption{Adapter hyperparameters. For contrastive adapters, batch size refers to the size of each minibatch sampled for updating with cross-entropy loss.}
\label{table:adapter_hparams}
\centering
\begin{adjustbox}{width=\textwidth}
{
\begin{tabular}{@{}lccccccc@{}}
\toprule
Dataset             & Max Epochs & Learning Rate & Weight Decay & Momentum & Batch Size & Hidden Dimension & Temperature \\ \midrule
Waterbirds          & 100        & 1e-3          & 5e-5         & 0.9      & 128        & 128              & 0.01                       \\
CelebA              & 50         & 1e-3          & 5e-5         & 0.9      & 128        & 128              & 0.01                       \\
BREEDS Living-17    & 100        & 1e-3          & 5e-5         & 0.9      & 128        & 128              & 0.01                       \\
BREEDS Nonliving-26 & 100        & 1e-3          & 5e-5         & 0.9      & 128        & 128              & 0.01                       \\
CIFAR-10.001         & 100        & 1e-3          & 5e-5         & 0.9      & 128        & 128              & 0.01                       \\
CIFAR-10.02         & 100        & 1e-3          & 5e-5         & 0.9      & 128        & 128              & 0.01                       \\
FMoW-WILDS          & 100        & 1e-3          & 5e-5         & 0.9      & 128        & 512              & 0.01                       \\
Amazon-WILDS        & 100        & 1e-3          & 5e-5         & 0.9      & 16         & 512              & 0.01                       \\
CivilComments-WILDS & 100        & 1e-3          & 5e-5         & 0.9      & 16         & 512              & 0.01                       \\ \bottomrule
\end{tabular}
}
\end{adjustbox}
\end{table}

\begin{table}[!h]
\caption{Specific contrastive adapter hyperparameters.}
\label{table:contrastive_adapter_hparams}
\centering
\begin{adjustbox}{width=\textwidth}
{
\begin{tabular}{@{}lcccc@{}}
\toprule
Dataset             & Number Positives & Number Negatives & Number Nearest Neighbors & Contrastive Temperature \\ \midrule
Waterbirds          & 2048             & 2048             & 4096                     & 0.1                     \\
CelebA              & 2048             & 2048             & 4096                     & 0.1                     \\
BREEDS Living-17    & 2048             & 2048             & 4096                     & 0.1                     \\
BREEDS Nonliving-26 & 512              & 512              & 1024                     & 0.1                     \\
CIFAR-10.001         & 512              & 512              & 1024                     & 0.1                     \\
CIFAR-10.02         & 512              & 512              & 1024                     & 0.1                     \\
FMoW-WILDS          & 2048             & 2048             & 2146                     & 0.1                     \\
Amazon-WILDS        & 2048             & 2048             & 2146                     & 0.1                     \\
CivilComments-WILDS & 2048             & 2048             & 2146                     & 0.1                     \\ \bottomrule
\end{tabular}
}
\end{adjustbox}
\vspace{-0.5cm}
\end{table}



\subsection{Data splits}
\label{appendix:experimental_details_data_splits}
We use the same train, validation, and test splits for Waterbirds, CelebA, FMoW-WILDS, Amazon-WILDS, and CivilComments-WILDS as in prior work. For BREEDS and CIFAR datasets that we adapt for our problem setting, we construct test splits by combining official test splits from the original benchmarks. We then create training and validation splits by combining the rest of the data from these benchmarks, and randomly splitting this into 80\% training data and 20\% validation data. No original test data is seen during training on our splits.

\subsection{Additional dataset assets details and discussion}
\label{appendix:experimental_details_assets}
\header{Dataset licenses} To curate CIFAR-10.0001 we use the CIFAR-10.1 dataset, which is distributed under the MIT License. The FMoW-WILDS dataset is distributed under the FMoW Challenge Public License\footnote{\href{https://github.com/fMoW/dataset/blob/master/LICENSE}{https://github.com/fMoW/dataset/blob/master/LICENSE}}. The CivilComments-WILDS dataset is distributed under CC0 1.0. The Amazon-WILDS dataset does not have a license, but is requested to be used for research purposes only~\cite{koh2021wilds}. We were not able to find explicit license information for CIFAR-10.2, Waterbirds, CelebA, or the BREEDS datasets. We note that the BREEDS datasets are sourced from ImageNet, which is distributed under the BSD 3-Clause License, and set up with code from the MadryLab robustness GitHub repository\footnote{\href{https://github.com/MadryLab/robustness}{https://github.com/MadryLab/robustness}}, which is distributed under a MIT license. The authors of the CelebA dataset provide a list of agreements\footnote{\href{http://mmlab.ie.cuhk.edu.hk/projects/CelebA.html}{http://mmlab.ie.cuhk.edu.hk/projects/CelebA.html}}, including that the dataset is used only for non-commercial research purposes.  

\header{Existing assets personally identifiable information and offensive content} The CelebA dataset consists of images of celebrity faces, which are personally identifiable. The dataset is also categorized by male and female identification at the time of curation, which may be outdated. The CivilComments-WILDS contains text samples flagged as toxic by toxicity classifiers~\cite{koh2021wilds}, which contain potentially offensive content. Both datasets are existing assets, and personal identifiability for CelebA and offensive content for CivilComments-WILDS can be checked by inspecting the original data inputs (images and text comments).


\subsection{Compute and resources}
\label{appendix:experimental_details_compute}

All experiments were run on a machine with 14 CPU cores and a single NVIDIA Tesla P100 GPU. For training a contrastive adapter on top of CLIP ResNet-50 Waterbirds embeddings, this took approximately 30 minutes to run 100 epochs. Other than the numbers reported from their original publications in Table~\ref{table:sota_comparison}, we report all numbers from running experiments on the same machine.

\subsection{Class prompt templates}
\label{appendix:experimental_details_prompts}
In Table~\ref{table:class_prompts}, we list the templates used to generate class prompts for each dataset. As a reminder, for each provided class name in a dataset, we create a prompt by inserting the class name into the prompt template. We then encode this prompt with a foundation model text encoder to get class embeddings.

\begin{table}[!t]
\caption{Class prompt templates or example prompts}
\label{table:class_prompts}
\centering
\begin{adjustbox}{width=0.8\textwidth}
{\small
\begin{tabular}{@{}lcl@{}}
\toprule
Dataset             & Foundation Model & Prompt template / example of prompt                                   \\ \midrule
Waterbirds          & CLIP             & ``This is a picture of a \texttt{[class\_name]}.'' \\
                    & CLOOB            & ``a \texttt{[class\_name]}''                       \\ \midrule
CelebA              & CLIP             & ``A photo of a celebrity with blond hair.''                           \\
                    & CLOOB            & ``A photo of a celebrity with blond hair.''                           \\ \midrule
BREEDS Living-17    & CLIP             & ``This is a picture of a \texttt{[class\_name]}.'' \\
                    & CLOOB            & ``This is a picture of a \texttt{[class\_name]}.'' \\ \midrule
BREEDS Nonliving-26 & CLIP             & ``A photo of a \texttt{[class\_name]}.''           \\
                    & CLOOB            & ``a \texttt{[class\_name]}''                       \\ \midrule
CIFAR-10.001        & CLIP             & ``a \texttt{[class\_name]}''                       \\
                    & CLOOB            & ``a \texttt{[class\_name]}''                       \\ \midrule
CIFAR-10.02         & CLIP             & ``a \texttt{[class\_name]}''                       \\
                    & CLOOB            & ``a \texttt{[class\_name]}''                       \\ \midrule
FMoW-WILDS          & CLIP             & ``satellite view of the \texttt{[class\_name]}''   \\
                    & CLOOB            & ``aerial view of an \texttt{[class\_name]}''       \\ \midrule
Amazon-WILDS        & GPT-Neo          & ``Negative''   \\ \midrule
CivilComments-WILDS & GPT-Neo          & ``Not toxic'' \\ \bottomrule
\end{tabular}
}
\end{adjustbox}
\vspace{-0.5cm}
\end{table}

\section{Additional related work discussion}
\label{appendix:extended_related_work}

We provide additional discussion of related work and connections to our work below.

\header{Zero-shot classification with foundation models} Our work builds on a growing literature on applying foundation models, large pretrained models that can be applied to various downstream tasks. These models demonstrate exciting promise in their ability to achieve accurate downstream transfer \emph{without} any additional finetuning~\cite{brown2020language, radford2021learning, bommasani2021opportunities}. In particular we consider the zero-shot capabilities of pretrained vision-language foundation models. These models, such as CLIP~\cite{radford2021learning}, ALIGN~\cite{jia2021scaling}, and CLOOB~\cite{furst2021cloob} are trained on massive amounts of naturally paired image-text data, \eg{} Internet images and their corresponding captions. Consisting of an image encoder (usually a ResNet or Vision Transformer) and a text encoder (usually a Transformer), such foundation models are commonly trained to learn a shared image-text embedding space where embeddings of images are most similar to embeddings of their corresponding caption text. While these objectives have been shown to lead to powerful representations~\cite{zhang2020contrastive,desai2021virtex}, a crucial element for successful zero-shot classification is training data scale~\cite{radford2021learning}. However, added scale can also be a double-edged sword; when zero-shot classification still makes undesirable mistakes, standard ways to correct for these mistakes via retraining can become prohibitively expensive. We study one such motivating instance via group robustness, and provide a first-step solution towards improving group robustness efficiently.

\header{Robustness of foundation models} Prior works have studied the robustness of foundation model inference to natural distribution shifts. \citet{radford2021learning} show that zero-shot CLIP models can be more robust to out-of-distribution (OOD) shifts than prior ImageNet-trained models, measured via better generalization to various dataset-level distribution shifts on ImageNet classes. 
However, they also show that finetuning, or updating the original weights, of CLIP models on ImageNet can reduce this OOD robustness. \citet{kumar2022fine, wortsman2021robust} thus propose finetuning methods that improve downstream in-distribution accuracy while maintaining out-of-distribution robustness. \citet{kumar2022fine} specifically study the trade-off between linear probing and finetuning, finding that finetuning on downstream data can improve generalization on in-distribution data over linear probing but more substantially hurt performance OOD data than linear probing. They show theoretically and empirically that a two-step strategy of first linear probing then full fine-tuning can combine the performance boosts of both. \citet{wortsman2021robust} focus on the OOD trade-off presented by \citet{radford2021learning} between a finetuned foundation model and its pretrained zero-shot weights. They propose weight-space ensembling (WiSE-FT), which computes a weighted average of the finetuned and pretraiend foundation model parameters, and show that the resulting averaged parameters can in some instances achieve higher performance on both data distributions that the model was finetuned on and unseen OOD data than the initial finetuned and zero-shot or pretrained models. They show this effect with both full finetuning and training a linear probe. Unlike these works, we focus on foundation model robustness to group shifts that occur within a dataset. We also compare against the linear probe version of WiSE-FT, and find that training adapters can be advantageous for achieving higher group robustness on various datasets. 

Recently, other works also study foundation model learned spurious correlations and biases. \citet{singla2022core} show how various models (including CLIP models) may rely on spurious artifacts to classify ImageNet images. \citet{berg2022prompt} aim to debias CLIP image embeddings of human faces using extra metadata (textual concepts or attributes) that the embeddings should ignore. Our evaluation is complementary, noting poor group robustness across multiple types of data sources (objects, animals, human faces, text). We also provide a method that works without additional training metadata.


\header{Improving group robustness of deep learning models} Improving the group robustness of deep learning models is a common deep learning challenge, where models may learn biases during training that lead to poor performance on certain groups. This is a widespread issue presented in contexts ranging from algorithmic fairness to healthcare diagnosis~\cite{blodgett2016demographic,hashimoto2018fairness,buolamwini2018gender}. Several methods exist to improve group robustness. We compare against several recent approaches in Section~\ref{sec:accessible_results}. While one effective strategy to improve group robustness is to upweight the error of worst-performing group during training~\cite{sagawa2019distributionally, sagawa2020investigation}, training group labels may be impractical to obtain in practice~\cite{sohoni2020no, oakden2020hidden}. We thus consider robustness approaches which aim to work without training group labels. Several approaches involve training two models; one model is first trained with standard ERM to help infer groups, and another trained with a robust objective using these inferred groups. Just Train Twice (JTT)~\cite{liu2021just} treats samples that the first model misclassifies as inferred minority group samples to upweight. JTT then upweights these samples by a hyperparameter factor, and trains a second model with ERM on this upsampled data. Environment Inference for Invariant Learning (EIIL)~\cite{creager2021environment} infers groups by assigning samples to group under which the ERM model maximally violates an Invariant Risk Minimization~\cite{arjovsky2019invariant} principle. It then trains a robust model with Group DRO~\cite{sagawa2019distributionally} using the inferred groups, which dynamically upweights the worst-performing groups during training. Correct-N-Contrast (CNC)~\cite{zhang2022correct} instead identifies samples with the same class labels but different ERM model predictions, and trains a robust model by using a contrastive loss to learn similar representations between these samples. Spread Spurious Attribute (SSA)~\cite{nam2022spread} specifically trains the first model to predict groups using a small set of group labels, before using Group DRO to train a robust model. Contrastive Input Morphing (CIM)~\cite{taghanaki2021robust} trains a network to transform the input features of an image to better present class-specific information shared across groups. \citet{idrissi2021simple} suggest that simply changing the training data by subsampling large classes (SUBY) or balancing class sampling probabilities (RWY), then training with ERM, can also improve group robustness.

\section{Additional experimental results}
\label{appendix:additional_experimental_results}

\subsection{Extended main results}
\label{appendix:expanded_main_results}

In Table~\ref{table:expanded_main_results_image} and Table~\ref{table:expanded_main_results_text} we report group robustness results evaluating all methods discussed in Section~\ref{sec:main_results} on all group robustness benchmarks. Table~\ref{table:expanded_main_results_image} contains results for image datasets, using CLIP RN-50 embeddings. Table~\ref{table:expanded_main_results_text} contains results for text datasets, using GPT-Neo 1.3B embeddings. As in Table~\ref{table:main_results}, we report the worst-group and average accuracies, along with their gap. Higher worst-group accuracy and smaller accuracy gap are indicative of better group robustness. All results are computed over three random seeds, with mean and one standard deviation included (error bars deferred to here from the main paper). Compared to alternative methods, contrastive adapting consistently improves group robustness over zero-shot classification, and obtains highest worst-group accuracy and smallest accuracy gap on datasets where training adapters with ERM fails. On datasets where ERM-trained adapters achieve best group robustness, contrastive adapters are also competitive or closest to ERM-trained adapters among other robustness methods.

\subsection{Contrastive adapter ablations}
In this section, we ablate different training components of contrastive adapting. We show that the presented combination leads to best worst-group accuracy on the Waterbirds dataset.
We also study how the number of positives and negatives used in contrastive sampling affects performance, and find that models do seem to benefit from a greater number of samples.

\subsubsection{Training component ablations}
We study the importance of the contrastive and cross-entropy components in contrastive adapting. For evaluation, we use the Waterbirds dataset, and run ablations comparing adapters trained on top of CLIP embeddings with (i) no contrastive component (Eq.~\ref{eq:contrastive_adapter}), (ii) no cross-entropy component (Eq.~\ref{eq:cross_entropy_adapter}), or the default proposed approach. We evaluate across five different CLIP models. All other training procedures are kept consistent.

In Table~\ref{table:adapter_ablation}, we report worst-group accuracies. We find that both contrastive and cross-entropy components are necessary for best worst-group accuracy. The contrastive objective leads to a substantial improvement over just the resampled cross-entropy loss (+17.9 pp on average). However, we also note that without the cross-entropy objective to learn sample embeddings close to their ground-truth class embeddings, we observe high variance in classification accuracy. We improve +26.9 pp on average using both objectives compared to contrastive alone.


\subsubsection{Effect of contrastive batch size}
While one advantage of training adapters is that because we train on embeddings, the memory size of our data inputs during training is much smaller than the traditional alternative (\eg{} storing an tensorized image). We can thus train with larger batch sizes. Here we study how contrastive batch size, \ie{} how many positives and negatives we sample per anchor, affects worst-group accuracy. On the Waterbirds and CelebA datasets and with CLIP RN-50 embeddings, we train a contrastive adapter with varying levels of positives and negatives. For both datasets, the default is 2048 positives and 2048 negatives per batch. We ablate these numbers with the following (positive, negative) combinations: (1, 1), (2, 2), (256, 256), (256, 512), (512, 256), (512, 512), (512, 1024), (1024, 512), (1024, 1024), (1024, 2048), (2048, 1024). 

\begin{figure}[t]
  \vspace{-0.25cm}
  \centering
  \includegraphics[width=0.8\textwidth]{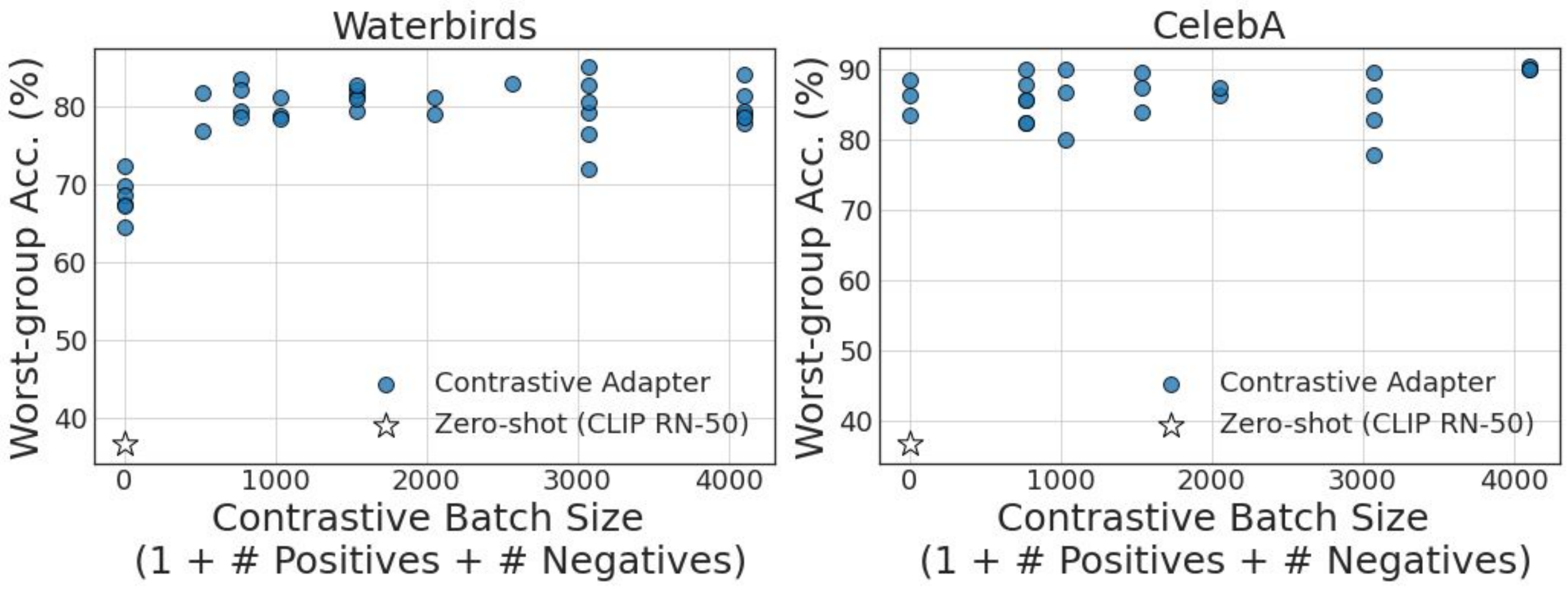}
  \caption{Effect of contrastive batch size on worst-group accuracy. With CLIP RN-50 embeddings, training contrastive adapters with larger batch sizes (greater number of positives and negatives) tends to help worst-group accuracy. However, even training with one positive and negative per batch leads to substantially greater worst-group accuracy than zero-shot classification.}
  \label{fig:batch_size_effect}
\end{figure}

In Figure~\ref{fig:batch_size_effect}, we plot the effect of smaller batch sizes on worst-group accuracy. We find that larger batch sizes weakly correspond to higher worst-group accuracy on both Waterbirds and CelebA. However, perhaps surprisingly, we still maintain a substantial improvement over zero-shot classification with just a single positive and negative per anchor. 


\begin{table}[t]
\caption{Contrastive adapter training component ablation. For five CLIP models, we report the worst-group accuracy (\%) on Waterbirds. Both contrastive and cross-entropy components are necessary for best worst-group accuracy. Without the cross-entropy objective to learn sample embeddings close to class embeddings (No cross-entropy), we observe high variance in classification accuracy.}
\label{table:adapter_ablation}
\centering
\begin{adjustbox}{width=1\textwidth}
{\small
\begin{tabular}{@{}lccccc@{}}
\toprule
Adapter Method Ablation  & RN-50          & RN-101          & ViT-B/32        & ViT-B/16       & ViT-L/14       \\ \midrule
No contrastive (Eq.~\ref{eq:contrastive_adapter})   & $56.3 \pm 1.5$ & $68.8 \pm 2.2$  & $56.7 \pm 2.4$  & $70.2 \pm 1.4$ & $75.1 \pm 1.0$ \\
No cross-entropy (Eq.~\ref{eq:cross_entropy_adapter}) & $60.7 \pm 8.3$ & $37.8 \pm 12.0$ & $23.1 \pm 10.5$ & $77.7 \pm 2.9$ & $82.4 \pm 2.0$ \\
Default          & $\mathbf{83.7 \pm 0.7}$ & $\mathbf{82.0 \pm 1.3}$  & $\mathbf{80.7 \pm 1.4}$  & $\mathbf{83.1 \pm 2.1}$ & $\mathbf{86.9 \pm 1.6}$ \\ \bottomrule
\end{tabular}
}
\end{adjustbox}
\end{table}

\subsection{Comparison to TIP-adapter and training sample nearest-neighbors lookup}

On representative benchmarks, we perform further comparison to the nearest-neighbor look-up approach employed by TIP Adapter~\cite{zhang2021tip}. Instead of learning transformed representations of pretrained embeddings, another approach to better classify a test sample is to use the class of its nearest training sample. 
Under the assumption that the training and test data are sampled from the same broader distribution and share the same groups, then test samples in a given group should embed closest to training samples in the same group. The training sample ground-truth class should then apply to the test sample.   
TIP adapter operates accordingly, keeping a cache of training sample embeddings available at test-time. One advantage is this allows for potentially more accurate classification \emph{without any} training. To test how well this idea fares for group robust classification, for each test sample we perform a look-up with \emph{all} training samples, using cosine similarity to identify nearest neighbors.

In Table~\ref{table:tip_adapter_comparison}, we compare TIP-adapter with zero-shot classification and contrastive adapting on the Waterbirds, CelebA, BREEDS Living-17, and CIFAR-10.02 group robustness benchmarks. For all methods, we use CLIP RN-50 pretrained embeddings. We find that TIP adapter improves worst-group accuracy over zero-shot classification on 3 out of 4 datasets, and notably achieves best worst-group accuracy on BREEDS Living-17 without training any additional parameters. However, the improvements are more marginal on Waterbirds and CIFAR-10.02. Contrastive adapting still achieves 30.4 pp higher worst-group accuracy over TIP adapter on average. This may suggest that learning a nonlinear transformation of the pretrained embeddings can still be helpful for better ``presenting'' class-specific information to classify samples by.

\begin{table}[!t]
\caption{Group robustness comparison to nearest training sample look-up / TIP Adapter~\cite{zhang2021tip}. Across representative benchmarks, on average contrastive adapting achieves 30.4 pp higher worst-group accuracy than the nearest training sample look-up employed by TIP-adapter. This supports learning non-linear transformations of pretrained embeddings to better classify samples.}
\label{table:tip_adapter_comparison}
\centering
\begin{adjustbox}{width=1\textwidth}
{\small
\begin{tabular}{@{}lccbccbccbccb@{}}
\toprule
                    & \multicolumn{3}{c}{Waterbirds}      & \multicolumn{3}{c}{CelebA}          & \multicolumn{3}{c}{BREEDS Living-17} & \multicolumn{3}{c}{CIFAR-10.02}      \\ \cmidrule(l){2-4}\cmidrule(l){5-7}\cmidrule(l){8-10}\cmidrule(l){11-13} 
Acc (\%)            & WG            & Avg  & Gap          & WG            & Avg  & Gap          & WG            & Avg  & Gap           & WG            & Avg  & Gap           \\ \midrule
Zero-shot (ZS)      & 36.6          & 92.2 & 55.6         & 74.0          & 81.9 & 7.9          & 6.0           & 86.7 & 80.7          & 39.1          & 69.9 & 30.8          \\
TIP Adapter         & 39.9          & 93.9 & 54.0         & 19.4          & 91.1 & 71.7         & \textbf{64.0} & 90.7 & \textbf{26.7} & 51.5          & 75.4 & 23.9          \\
Contrastive Adapter & \textbf{83.7} & 89.4 & \textbf{5.7} & \textbf{90.0} & 90.7 & \textbf{0.7} & 62.0          & 90.9 & 28.9          & \textbf{60.7} & 80.9 & \textbf{20.2} \\ \bottomrule
\end{tabular}
}
\end{adjustbox}
\end{table}



\subsection{Evaluation with respect to weight-space ensembling trade-off}
To provide additional perspective on how different embedding-only methods trade-off worst-group and average accuracy, we compare how these methods perform with respect to the accuracy trade-off traced out by weight-space ensembles. \citet{wortsman2021robust} show an interesting phenomenon where simply taking a weighted average of a trained linear probe and the original foundation model (either over the weights, or the outputs) can result in a ``pareto frontier'' of accuracy metrics. They specifically show that a weight-space ensemble can often achieve better OOD performance without sacrificing too much IID performance. compared to a single linear probe. In this context, we see how this trade-off occurs over average accuracy and worst-group accuracy across our representative set of group-robustness datasets. We also evaluate how other approaches (ERM adapters, DFR~\cite{kirichenko2022last}, contrastive adapters) fare along this trade-off.

\begin{figure}[t]
  \centering
  \includegraphics[width=1\textwidth]{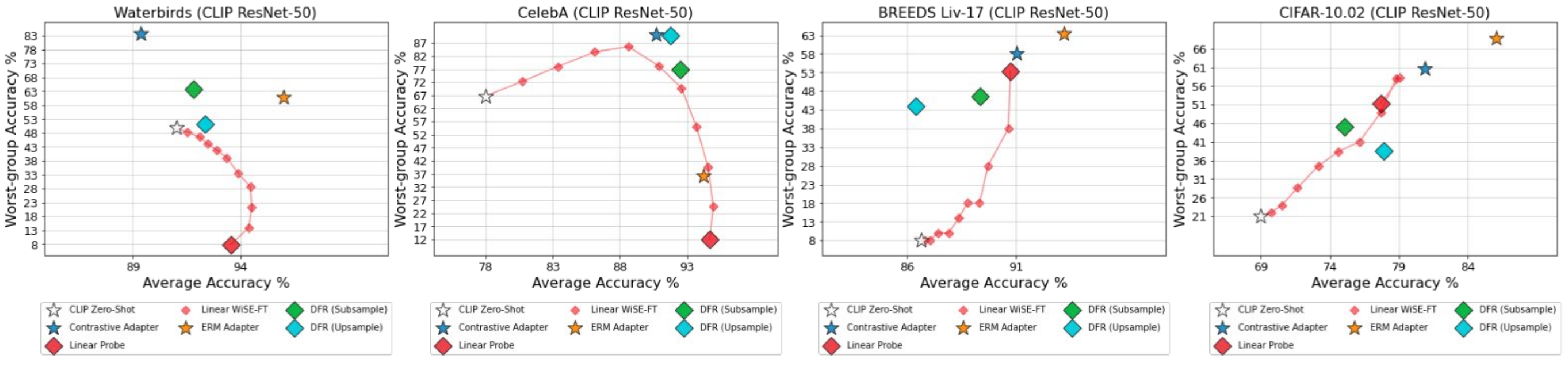}
  \vspace{-0.5cm}
  \caption{Plotting worst-group versus average accuracy trade-off against WiSE-FT ensemble (traced out) on representative datasets (Table~\ref{table:main_results}). Contrastive adapters (dark blue stars) consistently achieve higher worst-group accuracy than weight-space ensembles.}
  \label{fig:wise_ft_plot}
\end{figure}

In Figure~\ref{fig:wise_ft_plot}, we plot the accuracies of these methods run on CLIP RN-50 embeddings. We note several observations. Weight-space ensembles (WiSE-FT)  achieve the desired effect on certain datasets but not others. On CelebA and CIFAR-10.02, we find that an ensemble can obtain a better worst-group accuracy versus average accuracy trade-off than either zero-shot classification or linear probes. However, the single linear probe does at least as well as any ensemble in BREEDS Living-17, while the zero-shot classification does at least as well as any ensemble in Waterbirds. 

We also find that among other methods, contrastive adapting is the only evaluated approach that consistently achieves higher worst-group accuracy than any weight-space ensemble. While contrastive adapting places ``above'' the trade-off curve traced out by WiSE-FT on 3 out of 4 datasets (CelebA, BREEDS Living-17, and CIFAR-10.02), it tends to degrade average performance in favor of higher worst-group performance compared to other approaches. Further work can improve on how to raise worst-group performance without sacrificing any average performance when compared to zero-shot classification or ERM-trained adapters.

\begin{table}[!t]
\tabcolsep=0.1cm
\caption{Worst-group (WG) and average (Avg) accuracies (in \%) for zero-shot and efficient methods to improve CLIP RN-50 inference. \textbf{1st} / \uline{2nd} highest WG acc. and \textbf{1st} / \uline{2nd} smallest accuracy gap \textbf{bolded} / {\ul underlined} respectively. Contrastive adapter (Contrast. Adapter) bolded for visibility.}
\label{table:expanded_main_results_image}
\centering
\begin{adjustbox}{width=\textwidth}
\begin{tabular}{@{}lcccccccc@{}}
\toprule
\multicolumn{9}{c}{Waterbirds}                                                                                                                                                     \\
Acc.  & Zero-Shot      & Group Prompt         & ERM LP         & ERM Adapter             & WiSE-FT        & DFR (Sub)      & DFR (Up)             & \textbf{Contrast. Adapter} \\ \midrule
WG        & $49.8 \pm 0.0$ & $55.9 \pm 0.0$       & $7.9 \pm 1.0$  & $60.8 \pm 0.9$          & $49.8 \pm 0.0$ & $51.3 \pm 1.4$ & {\ul $63.9 \pm 1.5$} & $\mathbf{83.7 \pm 0.7}$    \\
Avg.      & $91.0 \pm 0.0$ & $87.8 \pm 0.0$       & $93.5 \pm 0.1$ & $96.0 \pm 0.1$          & $91.0 \pm 0.0$ & $92.4 \pm 0.1$ & $91.8 \pm 3.1$       & $89.4 \pm 0.9$             \\
\rowcolor{LightCyan} Gap       & 41.2           & 31.9                 & 85.6           & 35.2                    & 41.2           & 41.1           & {\ul 27.9}           & \textbf{5.7}               \\ \midrule
\multicolumn{9}{c}{CelebA}                                                                                                                                                         \\
Acc.      & Zero-Shot      & Group Prompt         & ERM LP         & ERM Adapter             & WiSE-FT        & DFR (Sub)      & DFR (Up)             & \textbf{Contrast. Adapter} \\ \midrule
WG        & $74.0 \pm 0.0$ & $70.8$               & $11.9 \pm 0.3$ & $36.1 \pm 1.4$          & $85.6 \pm 0.0$ & $76.9 \pm 1.4$ & {\ul $89.6 \pm 0.3$} & $\mathbf{90.0 \pm 0.4}$    \\
Avg.      & $81.9 \pm 0.0$ & $82.6$               & $94.7 \pm 0.0$ & $94.2 \pm 0.2$          & $88.6 \pm 0.0$ & $92.5 \pm 0.2$ & $91.8 \pm 0.1$       & $90.7 \pm 0.0$             \\
\rowcolor{LightCyan} Gap       & 7.9            & 11.8                 & 82.8           & 58.1                    & 3.0            & 15.6           & {\ul 2.2}            & \textbf{0.7}               \\ \midrule
\multicolumn{9}{c}{BREEDS Living-17}                                                                                                                                               \\
Acc.      & Zero-Shot      & Group Prompt         & ERM LP         & ERM Adapter             & WiSE-FT        & DFR (Sub)      & DFR (Up)             & \textbf{Contrast. Adapter} \\ \midrule
WG        & $6.0 \pm 0.0$  & $30.0 \pm 0.0$       & $53.3 \pm 0.9$ & $\mathbf{70.7 \pm 0.9}$ & $53.3 \pm 0.9$ & $46.7 \pm 3.4$ & $44.0 \pm 0.0$       & {\ul $62.0 \pm 1.6$}       \\
Avg       & $86.7 \pm 0.0$ & $90.6 \pm 0.0$       & $90.8 \pm 0.0$ & $93.9 \pm 0.1$          & $90.8 \pm 0.0$ & $89.3 \pm 0.3$ & $86.4 \pm 0.0$       & $90.9 \pm 0.3$             \\
\rowcolor{LightCyan} Gap       & 80.7           & 60.6                 & 37.5           & \textbf{23.2}           & 37.5           & 42.6           & 42.4                 & {\ul 28.9}                 \\ \midrule
\multicolumn{9}{c}{BREEDS Nonliving-26}                                                                                                                                            \\
Acc.      & Zero-Shot      & Group Prompt         & ERM LP         & ERM Adapter             & WiSE-FT        & DFR (Sub)      & DFR (Up)             & \textbf{Contrast. Adapter} \\ \midrule
WG        & $6.0 \pm 0.0$  & {\ul $56.0 \pm 0.0$} & $32.0 \pm 0.0$ & $\mathbf{61.3 \pm 1.9}$ & $36.7 \pm 0.9$ & $29.3 \pm 1.9$ & $30.0 \pm 4.1$       & $55.3 \pm 4.2$             \\
Avg       & $72.3 \pm 0.0$ & $87.1 \pm 0.0$       & $82.3 \pm 0.1$ & $92.1 \pm 0.2$          & $83.6 \pm 0.1$ & $80.6 \pm 0.1$ & $83.6 \pm 0.0$       & $88.1 \pm 0.6$             \\
\rowcolor{LightCyan} Gap       & 66.3           & {\ul 31.1}           & 50.3           & \textbf{30.8}           & 46.9           & 51.3           & 53.6                 & 32.8                       \\ \midrule
\multicolumn{9}{c}{CIFAR-10.001}                                                                                                                                                   \\
Acc.      & Zero-Shot      & Group Prompt         & ERM LP         & ERM Adapter             & WiSE-FT        & DFR (Sub)      & DFR (Up)             & \textbf{Contrast. Adapter}          \\ \midrule
WG        & $31.4 \pm 0.0$ & N/A                  & $44.0 \pm 1.4$ & $\mathbf{68.2 \pm 3.5}$ & $53.3 \pm 0.0$ & $18.1 \pm 4.3$ & $45.0 \pm 1.6$       & {\ul $59.7 \pm 4.1$}       \\
Avg       & $69.8 \pm 0.0$ & N/A                  & $75.2 \pm 0.2$ & $87.3 \pm 0.3$          & $81.1 \pm 0.0$ & $58.7 \pm 1.7$ & $78.3 \pm 0.1$       & $82.0 \pm 0.1$             \\
\rowcolor{LightCyan} Gap       & 38.4           & N/A                  & 31.2           & \textbf{19.1}           & 27.8           & 40.6           & 33.3                 & {\ul 22.3}                 \\ \midrule
\multicolumn{9}{c}{CIFAR-10.02}                                                                                                                                                    \\
Acc.      & Zero-Shot      & Group Prompt         & ERM LP         & ERM Adapter             & WiSE-FT        & DFR (Sub)      & DFR (Up)             & \textbf{Contrast. Adapter}          \\ \midrule
WG        & $39.1 \pm 0.0$ & N/A                  & $51.3 \pm 0.2$ & $\mathbf{68.8 \pm 0.5}$ & $58.2 \pm 0.2$ & $45.0 \pm 0.8$ & $38.5 \pm 2.1$       & {\ul $60.7 \pm 1.7$}       \\
Avg       & $69.9 \pm 0.0$ & N/A                  & $77.7 \pm 0.1$ & $86.0 \pm 0.5$          & $79.1 \pm 0.0$ & $75.0 \pm 0.3$ & $77.9 \pm 0.5$       & $80.9 \pm 0.2$             \\
\rowcolor{LightCyan} Gap       & 48             & N/A                  & 26.4           & \textbf{17.2}           & 20.9           & 30.0           & 39.4                 & {\ul 20.2}                 \\ \midrule
\multicolumn{9}{c}{FMoW-WILDS}                                                                                                                                                     \\
Acc.      & Zero-shot      & Group Prompt         & ERM LP         & ERM Adapter             & WiSE-FT        & DFR (Sub)      & DFR (Up)             & \textbf{Contrast. Adapter}          \\ \midrule
WG        & $10.5 \pm 0.0$ & -                    & $21.6 \pm 0.1$ & $\mathbf{41.3 \pm 0.5}$ & $21.6 \pm 0.1$ & $6.8 \pm 0.6$  & $27.0 \pm 0.2$       & {\ul $39.2 \pm 0.7$}       \\
Avg       & $13.2 \pm 0.0$ & -                    & $24.1 \pm 0.1$ & $43.6 \pm 0.5$          & $24.1 \pm 0.1$ & $10.2 \pm 0.5$ & $28.7 \pm 0.2$       & $41.9 \pm 0.1$             \\
\rowcolor{LightCyan} Gap       & 2.7            & -                    & 2.5            & {\ul 2.3}               & 2.5            & 3.4            & \textbf{1.7}         & 2.7                        \\ \bottomrule
\end{tabular}
\end{adjustbox}
\vspace{0.25cm}

\tabcolsep=0.1cm
\caption{Worst-group (WG) and average (Avg) accuracies (in \%) for zero-shot and efficient methods to improve GPT-Neo 1.3B inference. \textbf{1st} / \uline{2nd} highest WG acc. and \textbf{1st} / \uline{2nd} smallest accuracy gap \textbf{bolded} / {\ul underlined} respectively. Contrastive adapter (Contrast. Adapter) bolded for visability.}
\label{table:expanded_main_results_text}
\centering
\begin{adjustbox}{width=\textwidth}
\begin{tabular}{@{}lcccccccc@{}}
\toprule
\multicolumn{9}{c}{Amazon-WILDS}                                                                                                                               \\
Acc. & Zero-shot      & Group Prompt & ERM LP          & ERM Adapter    & WiSE-FT         & DFR (Sub)      & DFR (Up)             & \textbf{Contrast. Adapter} \\ \midrule
WG   & $79.4 \pm 0.0$ & N/A          & {\ul $87.2 \pm 0.3$}  & {\ul $87.2 \pm 0.3$} & {\ul $87.2 \pm 0.3$}  & {\ul $87.2 \pm 0.3$} & $85.4 \pm 0.8$       & $\mathbf{87.9 \pm 1.1}$    \\
Avg  & $86.7 \pm 0.0$ & N/A          & $93.3 \pm 0.2$  & $93.6 \pm 0.1$ & $93.3 \pm 0.2$  & $93.2 \pm 0.3$ & $92.7 \pm 0.7$       & $92.6 \pm 0.8$             \\
\rowcolor{LightCyan} Gap  & 7.3            & N/A          & 6.1             & 6.4            & 6.1             & {\ul 6.0}      & 7.3                  & \textbf{4.7}               \\ \midrule
\multicolumn{9}{c}{CivilComments-WILDS}                                                                                                                        \\
Acc. & Zero-shot      & Group Prompt & ERM LP          & ERM Adapter    & WiSE-FT         & DFR (Sub)      & DFR (Up)             & \textbf{Contrast. Adapter} \\ \midrule
WG   & $16.0 \pm 0.0$ & N/A          & $46.7 \pm 2.0$  & $32.1 \pm 1.5$ & $46.7 \pm 2.0$  & $47.4 \pm 0.9$ & {\ul $48.2 \pm 1.3$} & $\mathbf{50.1 \pm 1.5}$    \\
Avg  & $74.8 \pm 0.0$ & N/A          & $51.2 \pm 0.26$ & $37.7 \pm 0.7$ & $51.2 \pm 0.26$ & $51.9 \pm 0.8$ & $52.1 \pm 1.3$       & $54.2 \pm 0.5$             \\
\rowcolor{LightCyan} Gap  & 58.8           & N/A          & 4.5             & 5.6            & 4.5             & 4.5            &  \textbf{3.9}            & {\ul 4.1}               \\ \bottomrule
\end{tabular}
\end{adjustbox}
\end{table}

\end{document}